%% file: main.tex
\documentclass[10pt,twocolumn,letterpaper]{article}

\usepackage[pagenumbers]{iccv} % To force page numbers, e.g. for an arXiv version

\usepackage[accsupp]{axessibility}  % Improves PDF readability for those with disabilities.

\usepackage{colortbl}
\usepackage{multirow}
\usepackage{dsfont}
\newcommand{\colortable}{\cellcolor[RGB]{230,230,230}}
\usepackage{amssymb}% http://ctan.org/pkg/amssymb
\usepackage{pifont}% http://ctan.org/pkg/pifont
\newcommand{\cmark}{\ding{51}}%
\usepackage{makecell}

\usepackage{pgfplots}
\pgfplotsset{compat=1.18} 

\usepackage[labelfont=bf]{caption}
\usepackage{threeparttable}

\usepackage{placeins}

\usepackage{xcolor}
\definecolor{SteelBlue}{HTML}{4682B4}
\definecolor{TealGreen}{HTML}{3CB371}
\definecolor{Amber}{HTML}{FFBF00}
\definecolor{Terracotta}{HTML}{D2691E}

\definecolor{LightYellow}{RGB}{251,232,150}
\definecolor{DarkOrange}{RGB}{194,91,12}
\definecolor{SkyBlue}{RGB}{2,179,239}
\definecolor{LimeGreen}{RGB}{145,207,77}

\definecolor{Purple}{RGB}{198,132,231}
\definecolor{egopurple}{RGB}{128, 128, 255}


\definecolor{iccvblue}{rgb}{0.21,0.49,0.74}
\usepackage[pagebackref,breaklinks,colorlinks,allcolors=iccvblue]{hyperref}

\definecolor{brandeisblue}{rgb}{0.0, 0.44, 1.0}

\newcommand{\method}{{ST-Occ}\xspace}
\newcommand{\metric}{{STCV}\xspace}
\newcommand{\paradigm}{{unified}\xspace}
\newcommand{\Paradigm}{{Unified}\xspace}

\def\paperID{6503} % *** Enter the Paper ID here
\def\confName{ICCV}
\def\confYear{2025}

\title{Occupancy Learning with Spatiotemporal Memory}

\author{
Ziyang Leng$^{1}$ \hspace{0.5mm}
Jiawei Yang$^{2}$ \hspace{0.5mm}
Wenlong Yi$^{1}$ \hspace{0.5mm}
Bolei Zhou$^{1}$\\[1mm]
$^{1}$University of California, Los Angeles \hspace{0.5mm}
$^{2}$University of Southern California \hspace{0.5mm}
}

\begin{document}
\maketitle

\begin{abstract}
3D occupancy becomes a promising perception representation for autonomous driving to model the surrounding environment at a fine-grained scale. However, it remains challenging to efficiently aggregate 3D occupancy over time across multiple input frames due to the high processing cost and the uncertainty and dynamics of voxels.
To address this issue, we propose \method, a scene-level occupancy representation learning framework that effectively learns the spatiotemporal feature with temporal consistency. \method consists of two core designs: a spatiotemporal memory that captures comprehensive historical information and stores it efficiently through a scene-level representation and a memory attention that conditions the current occupancy representation on the spatiotemporal memory with a model of uncertainty and dynamic awareness. Our method significantly enhances the spatiotemporal representation learned for 3D occupancy prediction tasks by exploiting the temporal dependency between multi-frame inputs. Experiments show that our approach outperforms the state-of-the-art methods by a margin of 3 mIoU and reduces the temporal inconsistency by 29\%. The code and model are available at \url{https://github.com/matthew-leng/ST-Occ}.
\end{abstract}

\section{Introduction}
\label{sec:intro}

In recent years, vision-centric 3D occupancy representation has gained significant interest in autonomous driving \cite{li2023voxformer, tian2024occ3d, tong2023scene, li2023fb, yu2023flashocc, liu2023fully, cao2024pasco}. Closely related to Bird’s Eye View (BEV) representations, many prior efforts have sought to extend common BEV perception pipelines and techniques \cite{liu2023fully}—such as view transformation, decoder designs, and temporal fusion—to obtain high-quality 3D occupancy representations. 

Recent works leverage temporal information to improve the robustness of occupancy prediction \cite{tian2024occ3d, tong2023scene, li2023fb, wei2023surroundocc, ma2024cotr, yu2023flashocc}. To achieve this, historical features are typically stored on a frame-wise basis, aligned with the current frame, and processed in a recurrent \cite{li2022bevformer} or stacked manner \cite{huang2022bevdet4d}. However, with the extended height dimension in 3D occupancy representation, the memory and computation overheads of the temporal fusion process become a critical issue \cite{liu2023fully}. Moreover, occupancy prediction tasks require voxel-level detail, which demands higher granularity than the BEV representations for 3D detection tasks. Consequently, existing temporal fusion paradigms are often inefficient and insufficient in exploiting spatiotemporal information for 3D occupancy learning.

Much less has been explored on utilizing spatiotemporal information for occupancy representation learning. While high granularity in occupancy representation aids the prediction task \cite{tong2023scene}, the performance gains from temporal information integration remain limited \cite{li2023fb}. We attribute this to three main challenges: 1) Efficiency. The voxel-wise detail in occupancy representation makes it large and dense, so storing and processing multiple frames of historical features is resource-intensive, limiting the number of frames that can be used in temporal fusion. 2) Uncertainty. Due to factors like occlusion and varying lighting conditions \cite{li2024occmamba}, voxel-level uncertainties arise across frames, potentially accumulating noise and error during temporal fusion and negatively impacting prediction robustness and accuracy \cite{cao2024pasco}. 3) Dynamics. Dynamic instances in the scene introduce voxel shifts, resulting in misaligned historical features if not accurately modeled, which can hinder performance on dynamic instances.

\begin{figure*}[h]
    \centering
    % \fbox{\rule{0pt}{4cm}\rule{15cm}{0pt}} % Adjust size as needed
    \includegraphics[width=1.0\textwidth, trim=0 40 0 80, clip]{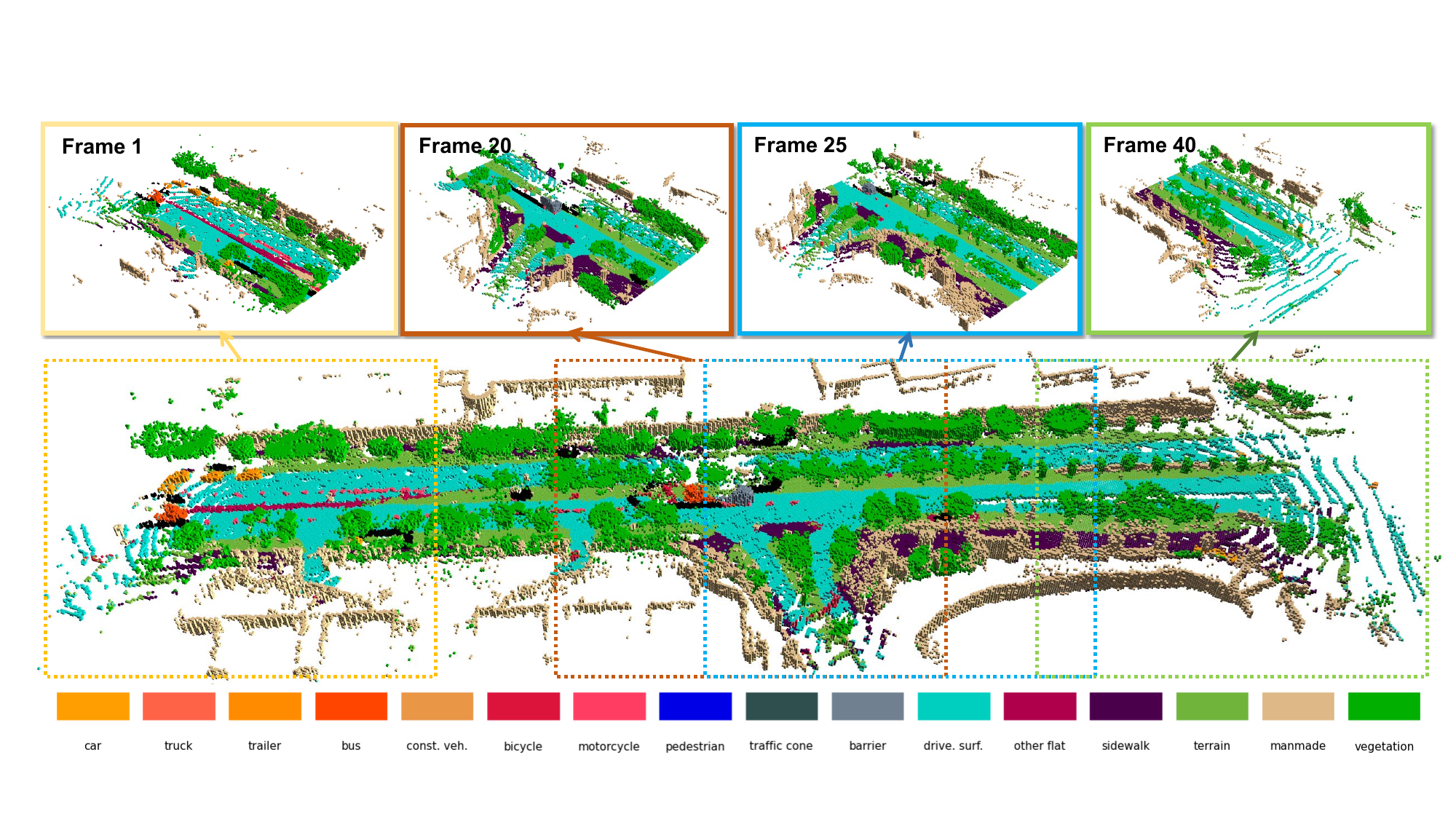}
    \caption{Occupancy prediction for a large-scale scene using our proposed \method. The first row shows ego-vehicle-centered predictions at different time steps and locations. The second row presents the scene-level occupancy prediction derived from our spatiotemporal memory, aggregating all 40 frames in the scene. The dashed rectangles are colored to correspond with their respective source frames above.}
    \label{fig:scene_demo}
\end{figure*}

To address the aforementioned problems, we propose constructing a spatiotemporal memory under scene-centered coordinates instead of ego vehicle-centered coordinates, demonstrated in \cref{fig:scene_demo}. In this way, we can not only store and process the historical feature efficiently in a recurrent way but also incorporate temporal clues that mitigate the uncertainty and compensate for the dynamics in occupancy representation. 
Therefore, we introduce a new paradigm: \paradigm temporal modeling, along with \method, a scene-level Spatiotemporal (ST) Occupancy representation learning framework designed to efficiently exploit the spatiotemporal information for the 3D occupancy prediction task. \method can perform 3D occupancy prediction in a streaming video approach \cite{park2022time} with a variable number of frames for temporal modeling, which makes it effective for constructing a complete and holistic representation of a large-scale scene with strong temporal consistency and robustness. Figure~\ref{fig:scene_demo} shows one occupancy prediction result by \method. 

\method consists of two core modules: a spatiotemporal memory bank and a memory attention. The spatiotemporal memory bank is constructed under scene-centered coordinates, which aim to capture comprehensive historical information, including historical representation, that assists the temporal modeling. The memory attention conditioned the current frame occupancy representation on the historical information from the spatiotemporal memory. This incorporates spatiotemporal information with uncertainty and dynamic awareness, which benefits the occupancy prediction. These two modules together compose the \paradigm temporal modeling for the framework.

Compared with the existing occupancy representation learning methods, the experiment shows the proposed \method surpasses the state-of-the-art method by a margin of 3 mIoU on the Occ3D benchmark \cite{tian2024occ3d} while maintaining computational and memory efficiency. Besides, our \paradigm temporal modeling is $2.8\times$ effective in utilizing temporal information. To further evaluate the temporal consistency of our framework's prediction, we design an evaluation metric to measure the occupancy prediction inconsistency between frames, and our method results in a 29\% decrease in temporal inconsistency. We summarize our contributions as follows:

\begin{itemize}
    \item We design a new \paradigm temporal modeling paradigm to achieve memory and computation-efficient temporal fusion.
    \item We propose a scene-level occupancy representation learning framework that implements the \paradigm temporal modeling. It exploits the spatiotemporal information with uncertainty and dynamic awareness.
    \item We conduct extensive experiments on the occupancy prediction task and our proposed temporal consistency evaluation metric, and our method outperforms state-of-the-art methods by substantial margins.
\end{itemize}

\section{Related Work}
\label{sec:related_work}

\paragraph{Camera-based 3D Occupancy Prediction.}
3D occupancy prediction aims to predict whether a voxel in the 3D space is occupied or not and its semantic class if occupied \cite{tian2024occ3d, tong2023scene, zheng2025occworld}. The use of occupancy maps can be traced back to robotics mapping and planning tasks \cite{dezert2015environment, schreiber2021dynamic}. The occupancy network introduced by Tesla \cite{tesla2021} brings the occupancy to autonomous driving perception. Recently, different camera-based 3D occupancy prediction works \cite{cao2022monoscene, tong2023scene, tian2024occ3d, wang2023openoccupancy, li2023fb} have been developed by extending vision-centric BEV perception pipelines \cite{li2022bevformer, huang2021bevdet, li2023fbbev}. For instance, OccNet \cite{tong2023scene} utilizes a cascade voxel decoder derived from the BEV decoder to reconstruct 3D occupancy. FB-OCC \cite{li2023fb} constructs the 3D features via forward-backward transformations used in FB-BEV \cite{li2023fbbev}. Another line of research aims to improve efficiency. For instance, FlashOcc \cite{yu2023flashocc} and COTR \cite{ma2024cotr} compress the intermediate representation to BEV or a smaller-scale representation. OctOcc \cite{ouyang2024octocc} and SparseOcc \cite{liu2023fully} design intermediate representations with varied levels of granularities to reduce the computational and memory cost, and they refine occupancy representations in a coarse-to-fine manner. These works contribute to obtaining a fine-grained representation with efficiency. 

Orthogonal to prediction precision, PasCo \cite{cao2024pasco} highlights the uncertainty awareness in occupancy prediction, which enables the model deployment in real-world safety-critical scenarios with noisy and ambiguous data. However, efforts to model the uncertainty and dynamics of occupancy from the perspective of temporal modeling remain less explored. 
If not modeled appropriately, these factors would lead to temporal inconsistency in the occupancy representation, resulting in decreased prediction performance with less robustness. 
Our work incorporates these factors into the temporal modeling process with efficiency, thus obtaining a fine-grained and robust representation for better prediction.

\paragraph{Temporal Modeling.}
Vision-centric perception significantly benefits from temporal modeling, which leverages cross-time information to obtain a better representation \cite{li2022bevformer, huang2022bevdet4d, park2022time, lin2022sparse4d, wang2023exploring, wang2024panoocc}. The historical information used in the temporal fusion helps the perception in the scenarios of occlusion, distortion, and lighting changes, among many others \cite{li2024occmamba}. Previous efforts mainly focus on temporal modeling for BEV perception. BEVFormer \cite{li2022bevformer} introduces temporal self-attention, which recurrently attends to previous ego-aligned BEV features. BEVDet and its follow-up \cite{huang2021bevdet, huang2022bevdet4d} differ in how to fuse past information: they align and concatenate historical features with present ones. Current occupancy prediction frameworks mostly extend these techniques from 2D BEV features to 3D occupancy features \cite{tian2024occ3d, tong2023scene, yu2023flashocc, li2023fb, shi2024occfiner}. However, the memory and computational costs brought by the dense occupancy representation are non-trivial and limit the scope of fused historical frames. Our work proposes a \paradigm temporal modeling, which is more memory and computation-efficient and effective in utilizing temporal information.

\section{Preliminaries}
\label{sec:preliminary}

We introduce the preliminaries and limitations of 3D occupancy representation and temporal fusion. We then discuss our \paradigm temporal modeling method to address these limitations at the end.

\paragraph{3D Occupancy Representation.}
\label{sec:3d_occ_repr}

3D occupancy representation is a world representation that offers a holistic view of the ego vehicle and its surroundings. Denoted as $\mathbf{V} \in \mathbb{R}^{H\times W\times Z\times C}$, an occupancy representation is a discretized 3D volumetric feature volume with a spatial shape of $H \times W \times Z$ and an additional feature dimension $C$, where each voxel contains scene features at its physical position \cite{huang2023tri}. 
With the additional height dimension compared with BEV representations, 3D occupancy representation usually contains more fine-grained geometric and semantic details of the scene and instances \cite{liu2023fully}. 
Despite its success, predicting occupancy frame by frame ignores the mutual information across time and is more vulnerable to sensor noise. Incorporating temporal information is thus an important component for robust occupancy learning. 

\begin{figure}[t]
    \centering
    \includegraphics[width=0.50\textwidth]{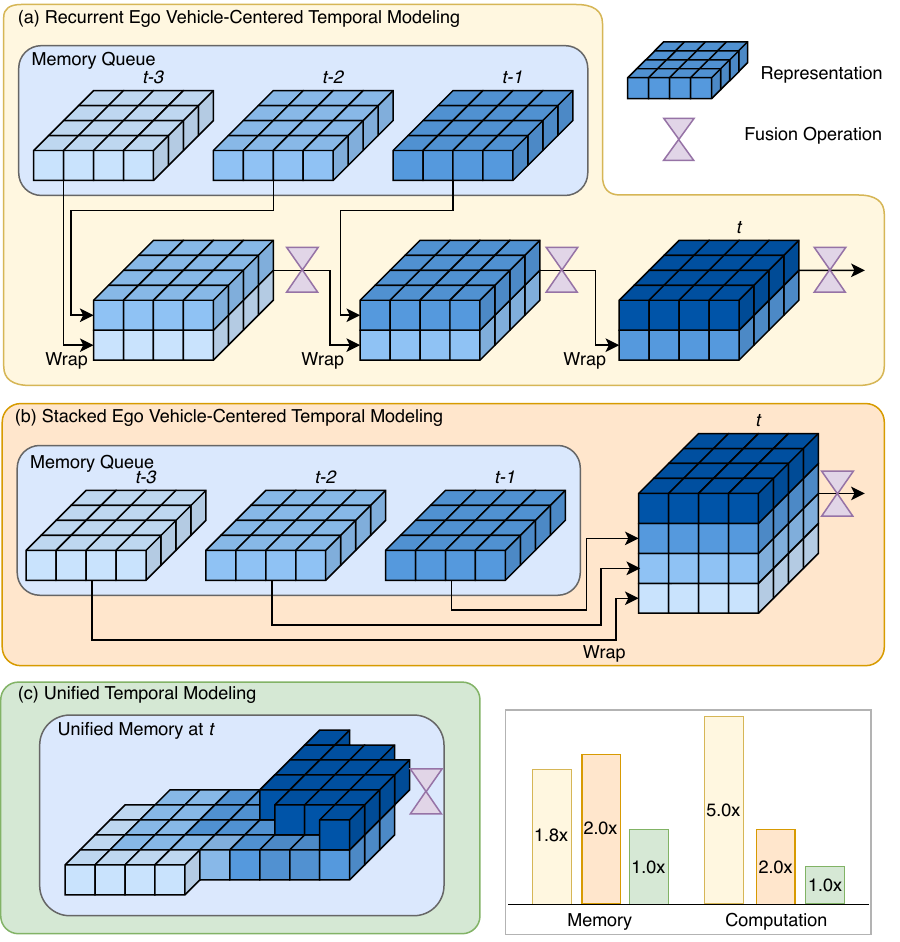}
    \vspace{-1em}
   \caption{Comparison of different temporal fusion paradigms, including (a) recurrent-based, (b) stacking-based ego vehicle-centered modeling, and (c) our proposed \paradigm temporal modeling. 
   Compared to previous approaches, our method requires significantly less memory and computation (multipliers are scaled relative to our approach).
   }
   \vspace{-1em}
   \label{fig:temporal_teaser}
\end{figure}

\paragraph{Queue-based Temporal Fusion.}
\label{sec:temp_fusion_paradigm}

Temporal fusion aims to use historical features to facilitate the perception in occluded or uncertain regions and dynamic instances.
Given a set of historical features $F = \{F^{t-k}, F^{t-k+1}, \dotso, F^{t} \}$ captured at different instantaneous timesteps from $t-k$ to $t$, we denote the temporal fusion process as 
\begin{equation}
    \tilde{F}_\text{out} = \psi (F),
\end{equation}
where  $\psi$ is the temporal fusion function, $k$ is number of historical frames used, $\tilde{F}_{\text{out}}$ is the output feature that incorporates temporal information.

Due to the ego vehicle motion, pose information $T_{t}$ is used to align the historical feature through the transformation matrix $T_{t-1}^t$, which is calculated as 
\begin{equation}
    T_{t-1}^t = T_t^{inv} \cdot T_{t-1}.
\end{equation}

Typical temporal fusion paradigms store historical frame features in a queue, align with the current frame, and process in a recurrent \cite{li2022bevformer} or stacked manner \cite{huang2022bevdet4d, li2023bevdepth}, shown in \cref{fig:temporal_teaser} (a) and (b), which can be expressed as
\begin{equation}
    \tilde{F}^t_\text{recurrent} = \psi(F^t, T_{t-1}^t\psi(F^{t-1}, T_{t-2}^{t-1}\psi(F^{t-2}, \dotso))),
\end{equation}
\begin{equation}
    \tilde{F}^t_\text{stack} = \psi(F^t, T_{t-1}^t \cdot F^{t-1}, T_{t-2}^{t} \cdot F^{t-2}, \dotso).
\end{equation}

The above approaches are widely adopted and effective in methods involving BEV representation. However, the extended height dimension considerably increases the storage and processing costs regarding occupancy representation.

\paragraph{\Paradigm Temporal Modeling}
\label{sec:scene-level_temp_model}

\Paradigm temporal modeling replaces the memory-heavy queue with a unified memory $M$ under scene-centered coordinates in \cref{fig:temporal_teaser} (c). The ego pose $T_t$ determines the region of interest (RoI) of timestamp $t$ in the unified memory. The process of incorporating temporal information is defined as:
\begin{equation}
    \tilde{F}^t_\text{\paradigm} = \psi(F^t \mid \chi[M_t, T_t]),
\end{equation}
where $\chi[\cdot]$ is the feature sampling operation and $M_t$ denotes the unified memory at $t$ timestamp.

The RoI in the unified memory is updated reversely using
\begin{equation}
    \chi[M_{t+1}, T_t] = \tilde{F}^t_\text{\paradigm}.
\end{equation}

\section{Method}
\label{sec:method}
 
\method employs a new \paradigm temporal modeling paradigm to exploit the spatiotemporal information in occupancy representation learning with uncertainty and dynamic awareness. We first introduce the pipeline of our proposed framework \method in (\S \ref{sec:framework}). We then talk about the two core modules (\S \ref{sec:global_memory}) and (\S \ref{sec:temp_fusion}) that realize the \paradigm temporal modeling. Lastly, we describe the temporal consistency evaluation (\S \ref{sec:temp_cons_metric}) and the loss functions (\S \ref{sec:optm}).

\subsection{\method}
\label{sec:framework}
\begin{figure*}[t]
    \centering
    \includegraphics[width=0.99\textwidth]{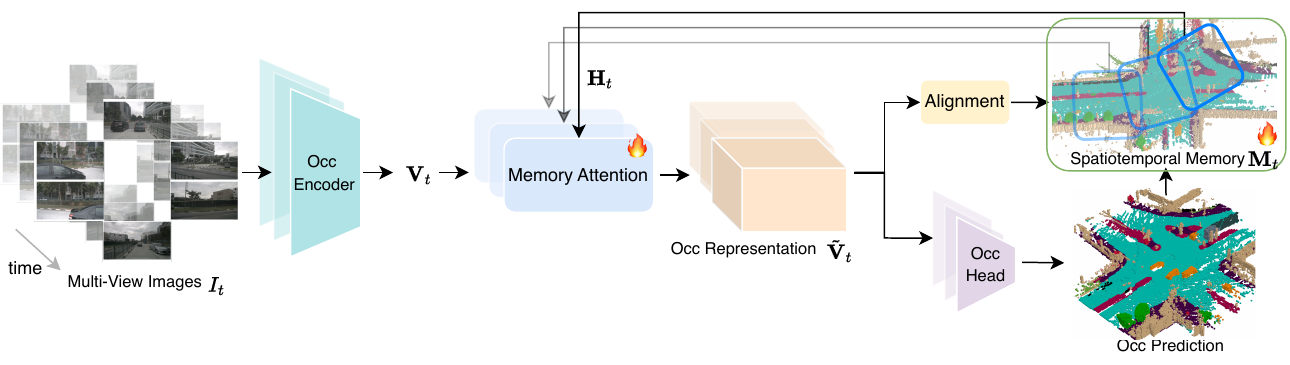}
    \vspace{-1em}
   \caption{
   \textbf{Overview of our \method}. \method implements the \paradigm temporal modeling using a spatiotemporal memory and a memory attention. The spatiotemporal memory captures comprehensive historical information in a scene-centered coordinate system, and the memory attention conditions the current occupancy representation on the spatiotemporal memory with uncertainty and dynamic awareness. 
   }
   \vspace{-1em}
   \label{fig:arch}
\end{figure*}

Our method aims to exploit the spatiotemporal information to learn the occupancy representation with strong temporal consistency and performance on occupancy prediction. As depicted in \cref{fig:arch}, \method contains two components for \paradigm temporal modeling: a spatiotemporal memory that preserves the temporal clues of historical input frames and a memory attention that conditions current frame occupancy representation on the spatiotemporal memory to incorporate historical information.

\cref{fig:arch} illustrates our pipeline. Specifically, given multi-view images input $I_t$ captured at timestamp $t$, the occupancy encoder extracts their ego vehicle-centered occupancy representations $\mathbf{V}_t$. Then, the memory attention conditions $\mathbf{V}_t$ on the historical information $\mathbf{H}_t$ to obtain the fused occupancy representation $\tilde{\mathbf{V}}_t$, where $\mathbf{H}_t$ is extracted from the corresponding region of interest (RoI) in the spatiotemporal memory. Lastly, we update this part of memory using the fused occupancy representation. Next, we will introduce the details of each component.

\subsection{Spatiotemporal Memory}
\label{sec:global_memory}

We design the spatiotemporal memory to efficiently store comprehensive historical information for temporal modeling of the occupancy representation learning. The spatiotemporal memory is constructed as a representation $\mathbf{M} \in \mathbb{R}^{H_\text{G} \times W_\text{G} \times Z_\text{G} \times C_\text{G}}$ at the beginning of each scene sequence. 
While the spatiotemporal memory representation is slightly larger than the ego vehicle-centered representation $\mathbf{V}$ due to ego motion, it is much more memory efficient when multiple temporal frames are used for temporal fusion.

When a total of $k$ temporal frames are used in the temporal modeling, the typical paradigms retain $k$ total representations while our \paradigm temporal modeling only requires one representation, thus more memory efficient.

We introduce not only the historical representation $\mathbf{V}$ but also other useful temporal attributes $\mu$ to preserve comprehensive information that would facilitate the temporal fusion process to the spatiotemporal memory. 

We define the temporal attributes of a voxel at position $\mathbf{p}$ as $\mu_\mathbf{p} \equiv \{\mathbf{c}_\mathbf{p}, \delta_\mathbf{p}, \mathbf{f}_\mathbf{p}\}$. For simplicity, we omit the subscript $\mathbf{p}$ from now on. Among these attributes, $\mathbf{c}\in\mathbb{R}^{N}$ is the historical class activation vector after the softmax operation, where $N$ is the number of classes and $\sum_{i=0}^{N-1}c_i = 1$; $\delta \in \mathbb{R}$ is the average log variance $\mathbf{s}$ over classes; and $\mathbf{f} \in \mathbb{R}^2$ is the occupancy flow vector in a top-down view.

The historical class activation of temporal attributes is updated using the class activation $\mathbf{c}_t$ from occupancy head, with an exponential decay of $\alpha$ follows
\begin{equation}
    \chi[\mathbf{M}_{t+1}\langle\mathbf{c}\rangle, T_t] = \text{softmax} \left(\alpha \mathbf{c}_t + (1 - \alpha) \chi[\mathbf{M}_t\langle\mathbf{c}\rangle, T_t] \right),
\label{eq:momentum}
\end{equation}
where the $\langle\cdot\rangle$ is the extraction operation from spatiotemporal memory.
We also add two additional networks in the occupancy head to predict log variance $\mathbf{s}$ and occupancy flow $\mathbf{f}$. 
They are updated to the spatiotemporal memory by
\begin{equation}
    \chi[\mathbf{M}_{t+1}\langle\delta\rangle, T_t] = \delta_t, \quad \chi[\mathbf{M}_{t+1}\langle \mathbf{f} \rangle, T_t] = \mathbf{f}_t.
\end{equation}

Finally, we update the historical representation using the memory attention conditioning
\begin{equation}
    \chi[\mathbf{M}_{t+1}\langle\mathbf{V}\rangle, T_t] = \tilde{\mathbf{V}}_t,
\end{equation}
which is introduced as follows.

\subsection{Memory Attention}
\label{sec:temp_fusion}

We design the memory attention to condition the initial occupancy representation $\mathbf{V}_t$ on the historical information $\mathbf{H}_t$ at timestamp $t$. The temporal information are incorporated into $\tilde{\mathbf{V}}_t$ follows
\begin{equation}
    \tilde{\mathbf{V}}_t = \psi(\mathbf{V}_t \mid \mathbf{H}_t) = \psi(\mathbf{V}_t \mid \chi[\mathbf{M}_t, T_t]),
\end{equation}
The historical information $\mathbf{H}_t$ consists of historical representation and temporal attributes. They are retrieved from the spatiotemporal memory using grid sampling.

To incorporate uncertainty and dynamic awareness in the memory attention, we use the uncertainty $u$ to balance the fused feature between historical and current representations, and we use the occupancy flow $\mathbf{f}$ to compensate the motion for voxels corresponding to dynamic instances.

To predict the uncertainty $u$, we use an MLP to encode temporal attributes as
\begin{equation}
    u = \text{MLP}(\mathbf{c}, \delta, \varepsilon),
\end{equation}
where $\varepsilon$ is the cosine similarity between current and historical representation at the same physical position.

\begin{figure}[t]
    \centering
    \includegraphics[width=0.50\textwidth]{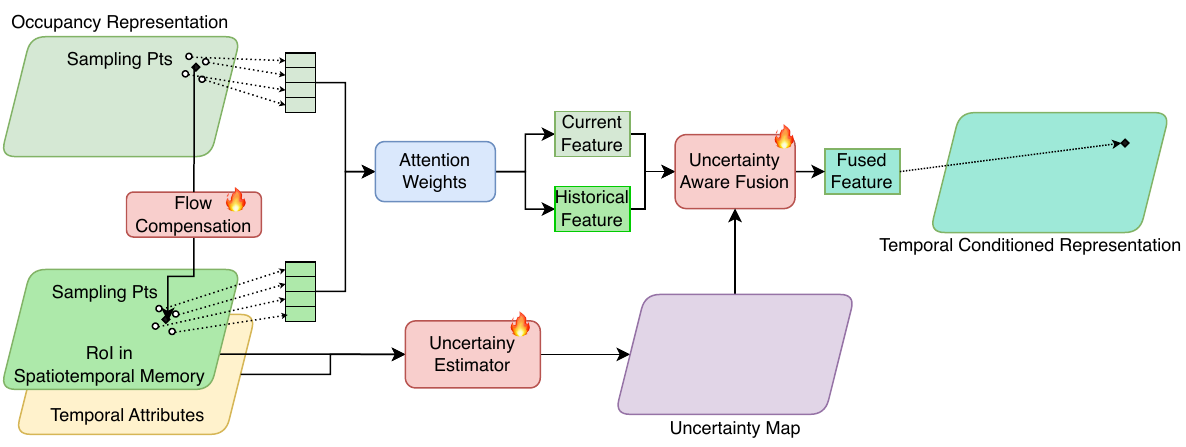}
   \caption{Illustration of memory attention with uncertainty and dynamic awareness.
   }
   \label{fig:temporal_fusion_module}
\end{figure}

The temporal conditioning process in our framework is built on the temporal self-attention (TSA) layer \cite{li2022bevformer}, with the encoded attributes incorporated to enable \textcolor{Purple}{uncertainty} and \textcolor{Terracotta}{dynamic} awareness. We use the initial occupancy representation as the \emph{query}, historical representation in RoI of spatiotemporal memory as \emph{key}, and \emph{value}. The process depicted in \cref{fig:temporal_fusion_module} uses deformable attention (DA) \cite{zhu2020deformable} which can be formulated as
\begin{equation}
    (1 - \textcolor{Purple}{u})\, \text{DA}(V_{t_p}, p + \textcolor{Terracotta}{f}, V_t)
+ \textcolor{Purple}{u}\, \text{DA}(V_{t_p}, p + \textcolor{Terracotta}{f}, \chi[\mathbf{M}_t, T_t]),
\end{equation}
where $V_{t_p}$ denotes the initial occupancy representation $V_t$ located at $p = (x, y, z)$. This design avoids feature misalignment and noise aggregation that causes temporal inconsistency while remaining entirely learnable.

\subsection{Measuring Temporal Consistency}
\label{sec:temp_cons_metric}

To assess the temporal consistency of occupancy predictions across frames in a sequence, we propose a new evaluation metric: mean Spatiotemporal Classification Variability (m\metric). This metric quantifies the classification variability of voxels representing the same real-world location over time, thereby measuring how stable occupancy predictions are across frames. 

\input{table/val}

To construct the correspondence of voxel between frames, we utilize our spatiotemporal memory to additionally store the historical occupancy prediction results $\mathbf{P}$ and occupancy ground truth $\mathbf{G}$ under scene-centered coordinates and update as follows
\begin{equation}
    \chi[\mathbf{M}_{t+1}\langle\mathbf{P}\rangle, T_t] = \mathbf{P}_t, \quad \chi[\mathbf{M}_{t+1}\langle \mathbf{G} \rangle, T_t] = \mathbf{G}_t.
\end{equation}

The \metric for timestamp $t$ is defined as
\begin{equation}
     \frac{\sum \mathds{1} [(\chi[\mathbf{M}_{t}\langle\mathbf{P}\rangle, T_t] \neq \mathbf{P}_t) \land (\chi[\mathbf{M}_{t}\langle\mathbf{P}\rangle, T_t] \neq \text{Free})]}{\sum \mathds{1}[\mathbf{P}_t \neq \text{Free}]},
\end{equation}
which calculates the percentage of classification changes in non-free voxels over the total number of non-free voxels. The m\metric is computed by averaging \metric across all frames:
\begin{equation}
    \text{m\metric} = \frac{1}{T} \sum_t^T \text{\metric}_t
\end{equation}

\subsection{Loss}
\label{sec:optm}

Our final loss function comprises three parts. 
The occupancy prediction loss $\mathcal{L}_{occ}$ follows the formulation in FB-OCC as
\begin{equation}
    \mathcal{L}_{occ} = \mathcal{L}_{fl} + \mathcal{L}_{ls} + \mathcal{L}^{geo}_{scal} + \mathcal{L}^{sem}_{scal} + \mathcal{L}_{d},
\end{equation}
which contains the Focal loss \cite{lin2017focal}, affinity loss following MonoScene \cite{cao2022monoscene}, Lovasz softmax loss \cite{berman2018lovasz}, and depth loss following FB-OCC \cite{li2023fb}.
The log variance prediction uses the Gaussian negative log-likelihood loss $\mathcal{L}_{nll}$ following
\begin{equation}
    \mathcal{L}_{nll} =  \sum_i \frac{1}{2} \left(\exp(-s_i)\| y_i -  \hat{y}_i \|^2 + s_i \right).
\end{equation}
Besides, an L1 loss $\mathcal{L}_{of}$ is used for the occupancy flow prediction. The training loss thus becomes
\begin{equation}
    \mathcal{L} = \mathcal{L}_{occ} + \mathcal{L}_{nll} + \mathcal{L}_{of}.
\end{equation}

\section{Experiments}
\label{sec:experiments}

\subsection{Dataset and Metrics}
We evaluate our method on the 3D occupancy benchmark, Occ3D \cite{tian2024occ3d}, constructed using the nuScenes dataset \cite{caesar2020nuscenes}. The benchmark consists of 1,000 driving sequences, each lasting 20 seconds, with RGB images captured from six cameras providing a 360-degree view. These sequences are divided into 700 training scenes, 150 validation scenes, and 150 test scenes. For each frame, the dataset includes 3D occupancy annotations within a range of [-40m, -40m, -1m, 40m, 40m, 5.4m] in ego-vehicle coordinates, with a voxel size of 0.4 meters. There are 18 voxel classes in total, one of which represents an unoccupied, free region.
Additionally, the Occ3D benchmark provides per-frame visibility masks, indicating whether each voxel is visible in the current camera view, to support both training and evaluation processes.

To assess the performance of our 3D occupancy prediction model, we use the mean Intersection-over-Union (mIoU) metric. Furthermore, we use our defined metric, m\metric, designed to evaluate the temporal consistency of occupancy predictions across consecutive frames.

\subsection{Implementation Details}

\paragraph{Network.} 

Our framework builds on the recent FB-OCC method \cite{li2023fb} and follows its experimental setup. We employ a ResNet50 \cite{he2016deep} backbone to extract perspective-view features from images of size $256 \times 704$. For more architectural details about the baseline, we refer readers to \cite{li2023fb}.
The output occupancy representation $\mathbf{V}$ is with dimensions $H = 100, W = 100, Z = 8$, and $C=80$. For fair comparison, we build on FB-OCC without its temporal fusion module to demonstrate the effectiveness of our \method. 

Our memory attention includes three temporal self-attention layers for temporal conditioning, along with a four-layer MLP to encode temporal attributes. The occupancy head includes three parallel three-layer convolutional networks to predict class activation $\mathbf{c}$, log variance $\mathbf{s}$, and occupancy flow $\mathbf{f}$.  
The decay factor of historical class activation (\textit{i.e.}, $\alpha$ in \cref{eq:momentum}) is set to 0.5.

\vspace{-1em}
\paragraph{Training.}

We train our \method with a learning rate of $2 \times 10^{-4}$ for 26 epochs.  Temporal modeling is excluded for the first three epochs to stabilize training. The ground truth for occupancy flow used in our flow prediction training is derived from nuScenes annotations in real-time by computing the instance bounding box offsets across time. Lastly, we utilize grid sampling function with bilinear interpolation to update our spatiotemporal memory.

\subsection{Results}

We conduct experiments to compare our proposed \method with previous state-of-the-art 3D occupancy prediction models, including TPVFormer \cite{huang2023tri}, OccFormer \cite{zhang2023occformer}, BEVFormer \cite{li2022bevformer}, CTF-Occ \cite{tian2024occ3d}, FB-OCC \cite{li2023fb}, and ViewFormer \cite{li2025viewformer}. We replace the temporal fusion modules of FB-OCC and ViewFormer with our framework while keeping other settings identical. \cref{tab:val} presents the 3D occupancy prediction results of all methods on the Occ3D benchmark \cite{tian2024occ3d}. Compared with the previous state-of-the-art FB-OCC, our method achieves a substantial improvement of 3 mIoU, with consistent performance gains across all classes. In examining the impact of temporal modeling, FB-OCC’s improvement with temporal fusion is 1.72 mIoU, while our approach achieves an increase of 4.74 mIoU—approximately $2.8 \times$ more effective than FB-OCC. 
Our framework also achieves 4.5 mIoU improvement on ViewFormer without temporal modeling, which is 25\% more effective in temporal modeling and surpasses the baseline by 0.9 mIoU. These results demonstrate the superior performance of our method and its effectiveness in temporal modeling.

We also compute our proposed temporal consistency metric, m\metric, for both our method and the baselines, as shown in \cref{tab:temporal_cons}. Compared with FB-OCC, our method reduces the temporal inconsistency by more than 25\%, whether applying the voxel visibility mask or not. These demonstrate the robustness and enhanced temporal consistency achieved by our method in occupancy prediction.

\input{table/temporal_consistency}

\begin{figure}[t]
    \centering
        \includegraphics[width=\linewidth]{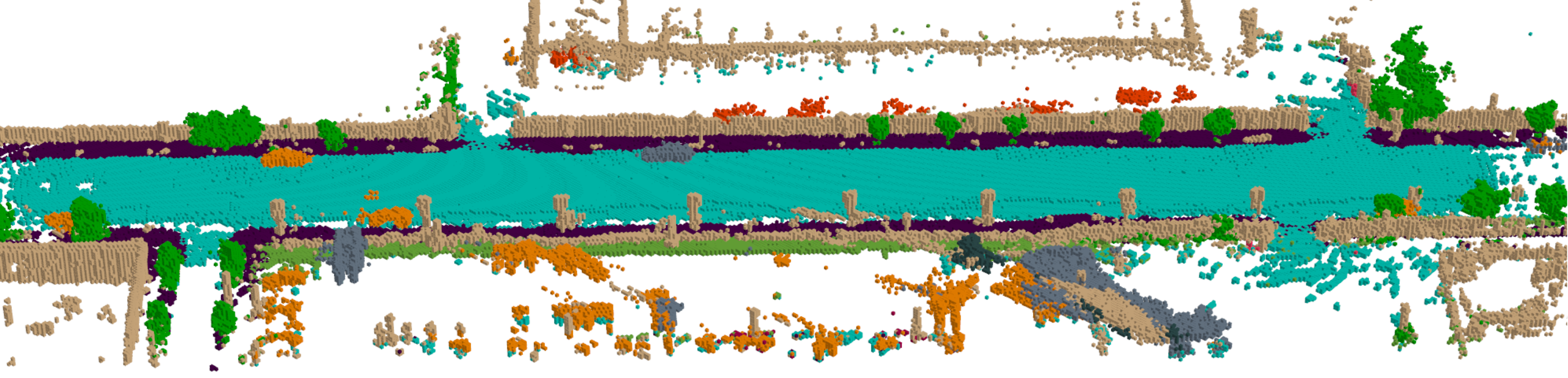}
    \vspace{-1.em}
    \caption{An example visualization of our aggregated spatiotemporal memory.}
    \label{fig:occ_pred_global}
    \vspace{-1.5em}
\end{figure}

\begin{figure}[t]
% \vspace{-2em}
    \centering
    \begin{subfigure}{0.23\textwidth}
        \centering
        % \fbox{\rule{0pt}{4cm}\rule{4cm}{0pt}}
        \includegraphics[width=\linewidth]{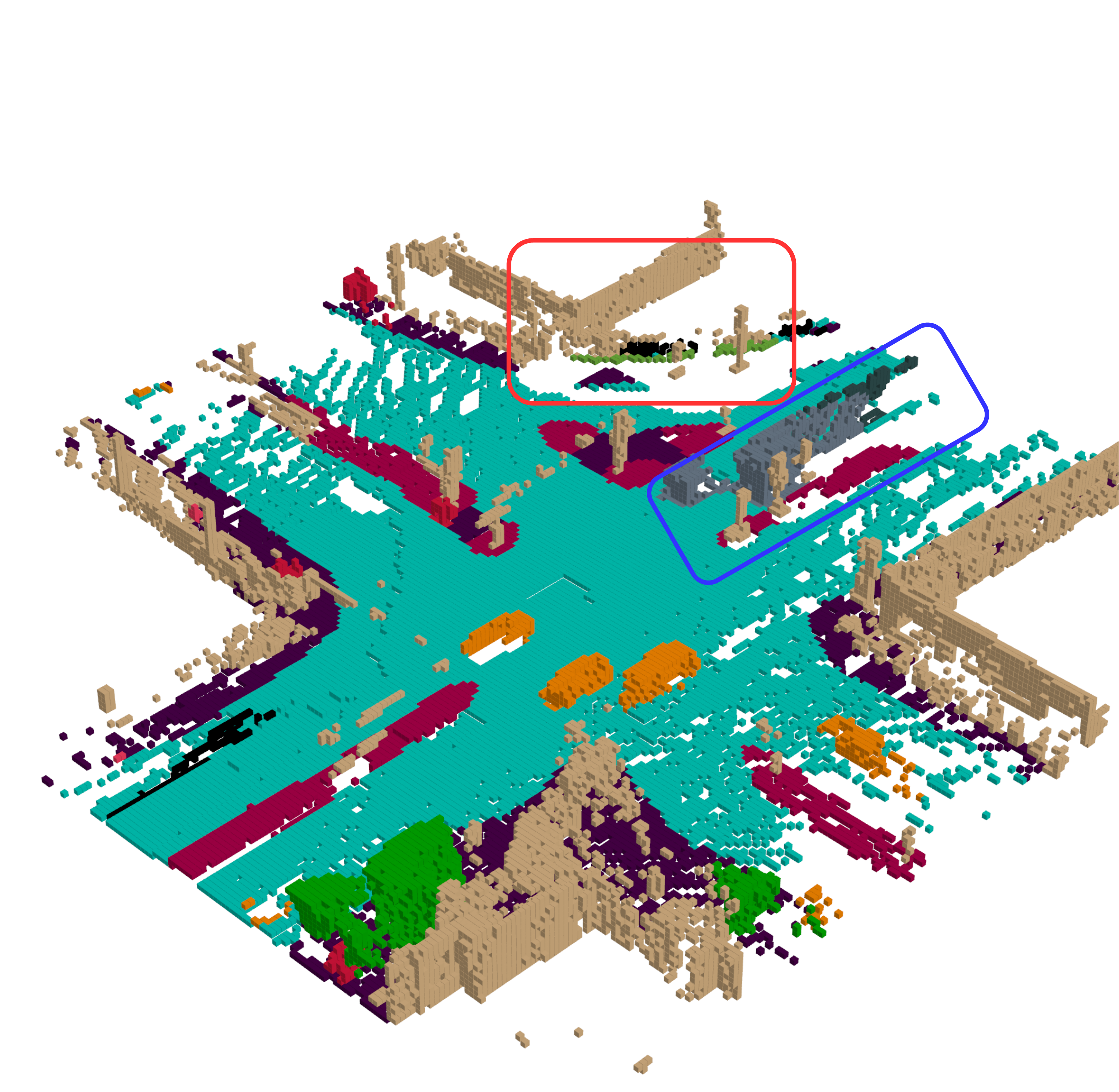}
        \caption{Prediction from FB-OCC}
        \label{fig:ori_pred}
    \end{subfigure}
    % \hfill
    \begin{subfigure}{0.23\textwidth}
        \centering
        % \fbox{\rule{0pt}{4cm}\rule{4cm}{0pt}}
        \includegraphics[width=\linewidth]{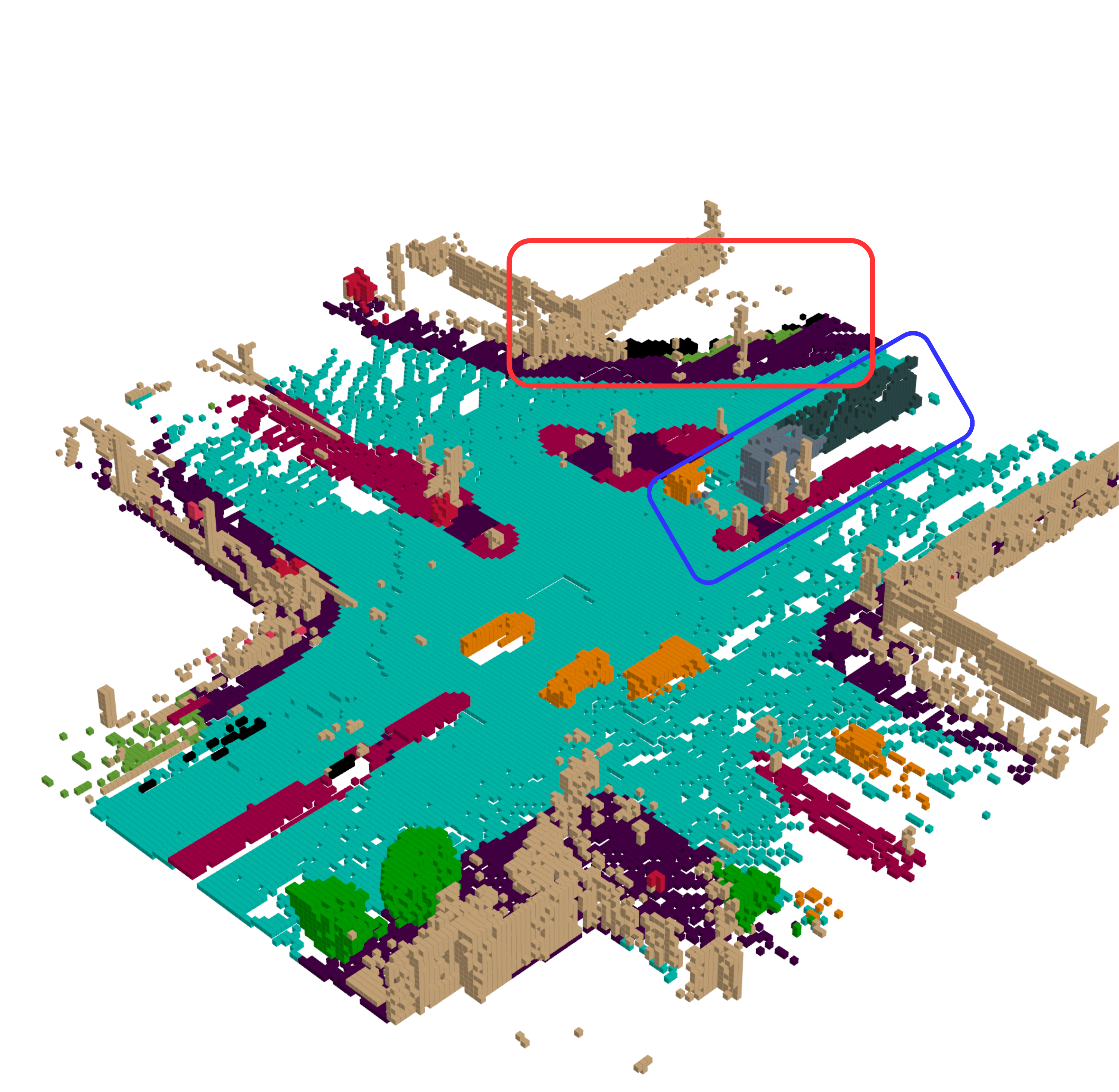}
        \caption{Prediction from \method}
        \label{fig:our_pred}
    \end{subfigure}
    \vspace{-1.em}
    \caption{A qualitative comparison on the occupancy prediction results between FB-OCC and \method.}
    \label{fig:occ_pred_cmpr}
    % \vspace{-1.5em}
\end{figure}

\paragraph{Qualitative Results} 

\cref{fig:scene_demo} presents an example of occupancy prediction results from our method. Our method produces high-quality occupancy predictions centered around the ego vehicle at various timestamps, displayed in the first row. The spatiotemporal memory retains the entire historical spatiotemporal representation, preserving a holistic view of the scene. The second row shows occupancy predictions derived from our spatiotemporal memory. \cref{fig:occ_pred_global} demonstrate our \method on another large-scale scene.

\cref{fig:occ_pred_cmpr} shows a qualitative comparison. Compared to FB-OCC, our approach more accurately predicts occupancy in some occluded regions, provides better instance classification and reduces noise in the predictions. These results demonstrate our method’s capability to model long-term temporal dependencies with both dynamic and uncertainty awareness, resulting in a more refined spatiotemporal representation and more robust occupancy prediction.

\subsection{Ablation Study}

\paragraph{Temporal Modeling Efficiency}

We compare the efficiency of our temporal modeling against a recurrent-based approach (VoxFormer) \cite{li2023voxformer} and a stacking-based approach (FB-OCC) \cite{li2023fb} in \cref{table:efficiency}. We use the same fusion operation, number of historical frames, and occupancy size for all methods to ensure fair comparison. These results indicate that our approach is the most efficient across metrics. \cref{tab:efficiency_plot} further demonstrates that our approach remains cost-efficient when incorporating more frames. 

\input{table/efficiency}
\input{table/efficiency_plot}

\paragraph{Uncertainty \& Dynamic Awareness}

\input{table/abl}

To understand our \method, we ablate different design choices in \cref{tab:abl}. The \textit{No Temporal} setting is identical to FB-OCC but without temporal fusion. The \textit{Mem. Attn.} corresponds to the memory attention in \cref{sec:temp_fusion}, with \textit{Dynamics} and \textit{Uncertainty} awareness optionally incorporated. Our method \method incorporates both uncertainty and dynamic awareness.

Compared with the model without temporal information, our vanilla memory attention achieves a performance increase of 3.78 mIoU, validating the effectiveness of our temporal fusion paradigm in exploiting historical information for occupancy prediction. Additionally, when augmented with our proposed dynamic awareness, our memory attention achieves around 1 IoU increase across classes covering dynamic instances. These improvements demonstrate our method's ability to capture instance dynamics. Further, incorporating uncertainty awareness yields a complementary performance boost, particularly in static classes, with approximately a 1 IoU increase. This result indicates that our approach can leverage uncertainty-aware attention to further mitigate inter-frame noise for better prediction in static regions. Lastly, equipping our memory attention with both uncertainty and dynamic awareness combines their strength and leads to the best performance across static and dynamic regions, increasing performance for 1 mIoU and consistently enhancing performance for all classes.

\input{table/full_abl}

\vspace{-1em}
\paragraph{Sub-components of \method}

\cref{tab:full_abl} presents an ablation study  on the sub-components of our \method, including 1) the use of historical class activation $\mathbf{c}$, feature similarity $\varepsilon$, and averaged log variance $\delta$ in uncertainty estimation. 2) occupancy flow $\mathbf{f}$, used to compensate for the movement of dynamic voxels across time. In line with our motivation, integrating more historical information into uncertainty estimation leads to more gains in performance, as the model can recurrently refine its predictions of uncertain regions using cross-time information.

\paragraph{Number of Frames for Temporal Fusion}

\input{table/frame_curve_sep}

In \cref{tab:frames}, we evaluate the impact of varying temporal fusion lengths on our method’s performance. Results show that our method benefits from long-term temporal fusion, achieving higher mIoU and reduced temporal inconsistency as the number of frames increases. Notably, compared to FB-OCC which uses 16 temporal frames, our method attains similar performance while only using 8 temporal frames, reducing temporal inconsistency by 20\% with an equal number of temporal frames during inference.

\section{Conclusion}
\label{sec:conclusion}

We propose \method, a scene-level occupancy representation learning framework. Our approach introduces a new temporal fusion paradigm, \paradigm temporal modeling, designed to capture long-term temporal dependencies in occupancy prediction. Our framework leverages spatiotemporal memory and memory attention, incorporating both uncertainty and dynamic awareness. Our \method achieves significant improvement over prior occupancy prediction methods. Consistent improvements on prediction precision (mIoU) and robustness (m\metric) from extensive experiments demonstrate the effectiveness of our \method.

There are limitations in our current approach that suggest potential directions for future work. Our method models dynamic voxels using an explicit estimator trained on existing nuScenes annotations. Future research could integrate this dynamic modeling directly into the temporal fusion process by deriving occupancy flow from temporal information, thereby reducing reliance on annotations. Another promising direction is to extend the \paradigm temporal modeling to sparse query-based perception methods.

\section*{Acknowledgements}
This work was supported by the NSF under Grants IIS-2339769 and CNS-2235012, and the Sony Focused Research Award.

{
    \small
    \bibliographystyle{ieeenat_fullname}
    \bibliography{main}
}

\input{suppl}

\end{document}

%% file: table/val.tex
\begin{table*}[t]
\small
\setlength{\tabcolsep}{3.5pt}
\centering
\resizebox{\textwidth}{!}{

\begin{tabular}{@{}l|c|>{\colortable}c|ccccccccccccccccc@{}}
\toprule
Method                  & Backbone  & mIoU  &\rotatebox{90}{others} &\rotatebox{90}{barrier} & \rotatebox{90}{bicycle} &  \rotatebox{90}{bus}   & \rotatebox{90}{car}   & \rotatebox{90}{const. veh.} & \rotatebox{90}{motorcycle} & \rotatebox{90}{pedestrian} & \rotatebox{90}{traffic cone} & \rotatebox{90}{trailer} & \rotatebox{90}{truck} & \rotatebox{90}{driv. surf.} & \rotatebox{90}{other flat} & \rotatebox{90}{sidewalk} & \rotatebox{90}{terrain} & \rotatebox{90}{manmade} & \rotatebox{90}{vegetation} \\
\midrule
TPVFormer  & \multirow{4}{*}{Res101} & 27.83 & 7.22 & 38.90 & 13.67 & 40.78 & 45.90 & 17.23 & 19.99 & 18.85 & 14.30 & 26.69 & 34.17 & 55.65 & 35.47 & 37.55 & 30.70 & 19.40 & 16.78 \\
OccFormer  &  & 21.93 & 5.94 & 30.29 & 12.32 & 34.40 & 39.17 & 14.44 & 16.45 & 17.22 & 9.27 & 13.90 & 26.36 & 50.99 & 30.96 & 34.66 & 22.73 & 6.76 & 6.97 \\
BEVFormer  &  & 26.88 & 5.85 & 37.83 & 17.87 & 40.44 & 42.43 & 7.36 & 23.88 & 21.81 & 20.98 & 22.38 & 30.70 & 55.35 & 28.36 & 36.00 & 28.06 & 20.04 & 17.69 \\
CTF-Occ  &  & 28.53 & 8.09 & 39.33 & 20.56 & 38.29 & 42.24 & 16.93 & 24.52 & 22.72 & 21.05 & 22.98 & 31.11 & 53.33 & 33.84 & 37.98 & 33.23 & 20.79 & 18.00 \\
\midrule

\textcolor{SteelBlue}{FB-OCC$^{\dagger}$} & \multirow{3}{*}{Res50} & 
\textcolor{SteelBlue}{37.39} & \textcolor{SteelBlue}{12.17} & \textcolor{SteelBlue}{44.83} & 
\textcolor{SteelBlue}{25.73} & \textcolor{SteelBlue}{42.61} & \textcolor{SteelBlue}{47.97} & 
\textcolor{SteelBlue}{23.16} & \textcolor{SteelBlue}{25.17} & \textcolor{SteelBlue}{25.77} & 
\textcolor{SteelBlue}{26.72} & \textcolor{SteelBlue}{31.31} & \textcolor{SteelBlue}{34.89} & 
\textcolor{SteelBlue}{78.83} & \textcolor{SteelBlue}{41.42} & \textcolor{SteelBlue}{49.06} & 
\textcolor{SteelBlue}{52.22} & \textcolor{SteelBlue}{39.07} & \textcolor{SteelBlue}{34.61} \\

FB-OCC &  & 39.11 & 13.57 & 44.74 & 27.01 & 45.41 & 49.10 & 25.15 & 26.33 & 27.86 & 27.79 & 32.28 & 36.75 & 80.07 & 42.76 & 51.18 & 55.13 & 42.19 & 37.53 \\
\rowcolor[RGB]{230,230,230} \method (ours) & & \textbf{42.13} & \textbf{14.36} & \textbf{49.62} & \textbf{27.77} & \textbf{46.28} & \textbf{52.55} & \textbf{26.87} & \textbf{29.79} & \textbf{29.83} & \textbf{31.39} & \textbf{35.40} & \textbf{39.03} & \textbf{84.26} & \textbf{47.72} & \textbf{56.09} & \textbf{59.85} & \textbf{45.27} & \textbf{40.11} \\

\midrule

\textcolor{SteelBlue}{ViewFormer$^{\dagger*}$} & \multirow{3}{*}{Res50} & 
\textcolor{SteelBlue}{37.80} & \textcolor{SteelBlue}{9.90} & \textcolor{SteelBlue}{44.89} & 
\textcolor{SteelBlue}{22.67} & \textcolor{SteelBlue}{42.84} & \textcolor{SteelBlue}{48.90} & 
\textcolor{SteelBlue}{21.39} & \textcolor{SteelBlue}{24.52} & \textcolor{SteelBlue}{25.22} & 
\textcolor{SteelBlue}{24.93} & \textcolor{SteelBlue}{29.18} & \textcolor{SteelBlue}{34.56} & 
\textcolor{SteelBlue}{81.93} & \textcolor{SteelBlue}{44.07} & \textcolor{SteelBlue}{53.72} & 
\textcolor{SteelBlue}{55.50} & \textcolor{SteelBlue}{42.18} & \textcolor{SteelBlue}{36.29} \\

ViewFormer$^*$ &   & 41.44 & 11.63 & 50.16 & 26.49 & 44.39 & \textbf{53.36} & 22.85 & 27.80 & 27.74 & 29.95 & 33.04 & 39.39 & 84.67 & 48.08 & 57.43 & 59.64 & 47.57 & 40.38 \\
\rowcolor[RGB]{230,230,230} ViewFormer$^{\dagger}$ + ST-Occ (ours) & & \textbf{42.30} & \textbf{12.00} & \textbf{50.61} & \textbf{27.93} & \textbf{45.81} & 53.24 & \textbf{24.79} & \textbf{29.18} & \textbf{28.63} & \textbf{30.93} & \textbf{33.53} & \textbf{39.58} & \textbf{85.28} & \textbf{49.42} & \textbf{58.39} & \textbf{60.39} & \textbf{48.02} & \textbf{41.42} \\
\bottomrule
\end{tabular}
}
\caption{3D occupancy prediction results on Occ3D benchmark. $^{\dagger}$ without temporal information. $^*$ reproduced using its official code.
The \emph{direct} baselines of our method are colored by \textcolor{SteelBlue}{steelblue} (Best viewed in color).
}
\vspace{-1em}
\label{tab:val}
\end{table*}

%% file: table/temporal_consistency.tex
\begin{table}[t]
\small
\centering
\resizebox{\linewidth}{!}{
\begin{tabular}{@{}l|cc}
\toprule
Method                            & m\metric (\%) $\downarrow$ & m\metric$^{\dagger}$ (\%) $\downarrow$ \\ 
\midrule
FB-OCC                             & 12.18                   & 8.57                    \\
Mem. Attn.                  & 9.25                   & 6.64                    \\
Mem. Attn. + Uncertainty & 8.85                    & 6.53                    \\
\rowcolor[RGB]{230,230,230} Mem. Attn. + Uncertainty + Dynamics           & \textbf{8.68}                    & \textbf{6.48}                   \\ 
\bottomrule
\end{tabular}
}
\vspace{-1em}
\caption{Temporal consistency evaluation results on Occ3D datasets of FB-OCC and \method under various settings. \textit{Mem. Attn.} denotes the memory attention. $^{\dagger}$ without applying voxel visibility mask. Our default setting is colored in grey.}
\vspace{-1em}
\label{tab:temporal_cons}
\end{table}

%% file: table/efficiency.tex
\begin{table}[t!]
    \centering
    \resizebox{1.0\linewidth}{!}{ 
    \begin{tabular}{l|cc|cc}
        \toprule
       \makecell{Temporal \\ Modeling} & \makecell{Training Mem.  \\ (GB) $\downarrow$} & \makecell{Fusion Time \\ (ms) $\downarrow$} & \makecell{Inference Mem. \\ (GB) $\downarrow$ }  & FPS$\uparrow$  \\ 
        \midrule
        Recurrent & 12.89 & 705 & 10.08  & 5.95   \\
        Stacked & 19.02 & 84 & 11.29 & 5.42   \\ 
        \rowcolor[RGB]{230,230,230} Unified (ours) & \textbf{10.90} & \textbf{24} & \textbf{5.57} & \textbf{8.65}  \\
        \bottomrule
    \end{tabular}
    }
    \vspace{-0.8em}
    \caption{Training and inference efficiency comparison between three temporal modeling paradigms. Our approach achieves the best efficiency across training and inference.}
    \label{table:efficiency}
\end{table}

%% file: table/efficiency_plot.tex
\begin{figure}[t!]
    \centering
    \vspace{-1em}
    % First subplot for mIoU
    \begin{tikzpicture}
    \hspace{-3em}
        \begin{axis}[
        legend style={nodes={scale=0.75, transform shape}},
        label style={font=\footnotesize},
        title style={font=\footnotesize},
            title=Training Fusion Time (ms),
            title style={yshift=-4pt},
            width=4.7cm,
            height=4cm,
            axis lines=left,
            xlabel={Number of Frames},
            ylabel style={yshift=-15pt},
            symbolic x coords={4, 8, 16, 40},
            xtick=data,
            % ymode=log, % Use logarithmic scale for the Y-axis
            % log basis y={2},
            % ytick={16,32,64,128,256,512,1024,2048},
            scaled y ticks=false, % Disable automatic scaling of Y-ticks
        ymin=0, ymax=11.6, % Adjust range of transformed Y values
        xtick=data,
        ytick={0, 1.16, 2.36, 3.74, 5.29, 7.48, 10.04, 11.23},
        yticklabels={0, 24, 100, 250, 500, 1000, 1800, 2000},
            legend style={at={(0.43,0.6)}, anchor=south, font=\small},
            label style={font=\small},
            tick label style={font=\small},
            extra y ticks={10},
            extra y tick labels={},
            extra y tick style={grid=major},
        ]
            % Plot mIoU
            \addplot[
                color=green,
                mark=*,
                mark options={scale=0.8},
                thick
            ] coordinates {
             (4, 2.44)
             (8, 4.354)
            (16, 6.48)
            (40, 10)
            };
            \addlegendentry{Recurrent}
            
            \addplot[
                color=cyan,
                mark=triangle*,
                mark options={scale=0.8},
                thick
            ] coordinates {
            (4, 1.26) 
            (8, 1.594)
            (16, 2.17) 
            (40, 3.52) 
            };
            \addlegendentry{Stacked}

            \addplot[
                color=magenta,
                mark=diamond*,
                mark options={scale=1.0},
                thick
            ] coordinates {
            (4, 1.14)
            (8, 1.14)
            (16, 1.14) 
            (40, 1.14) 
            };
            \addlegendentry{Unified (ours)}
        \end{axis}
    \end{tikzpicture}
    % Space between subplots
    % \hspace{0.05\columnwidth}
    % Second subplot for m\metric
    \hspace{-2.5em}
    \begin{tikzpicture}
        \begin{axis}[
        legend style={nodes={scale=0.75, transform shape}},
        label style={font=\footnotesize},
        title style={font=\footnotesize},
            title=Training Memory (GB),
            title style={yshift=-4pt},
            width=4.7cm,
            height=4cm,
            axis lines=left,
            xlabel={Number of Frames},
            ylabel style={yshift=-15pt},
            symbolic x coords={4, 8, 16, 40},
            xtick=data,
        ymin=5, ymax=45, % Adjust range of transformed Y values
        ytick={5, 10, 20, 30, 40}, 
            legend style={at={(0.42,0.6)}, anchor=south, font=\small},
            label style={font=\small},
            tick label style={font=\small},
            extra y ticks={42},
            extra y tick labels={},
            extra y tick style={grid=major},
        ]
            % Plot mIoU
            \addplot[
                color=green,
                mark=*,
                mark options={scale=0.8},
                thick
            ] coordinates {
                (4, 12.534)
                (8, 12.662)
            (16, 12.877)
            (40, 19.373)
            };
            \addlegendentry{Recurrent}
            
            \addplot[
                color=cyan,
                mark=triangle*,
                mark options={scale=0.8},
                thick
            ] coordinates {
                (4, 9.506)
                (8, 11.831)
            (16, 19.022) 
            (40, 42.477) 
            };
            \addlegendentry{Stacked}

            \addplot[
                color=magenta,
                mark=diamond*,
                mark options={scale=1.0},
                thick
            ] coordinates {
                (4, 10.899) 
                (8, 10.899)
            (16, 10.899) 
            (40, 10.899) 
            };
            \addlegendentry{Unified (ours)}
        \end{axis}
        % \hspace{6em}
    \end{tikzpicture}
    \hspace{-2em}
    \vspace{-1em}
    \caption{Effect of the number of temporal frames incorporated on the training efficiency of three temporal fusion paradigms. Our temporal modeling yields significantly less computational and memory footprint across number of frames.
}
    \vspace{-1em}
    \label{tab:efficiency_plot}
\end{figure}
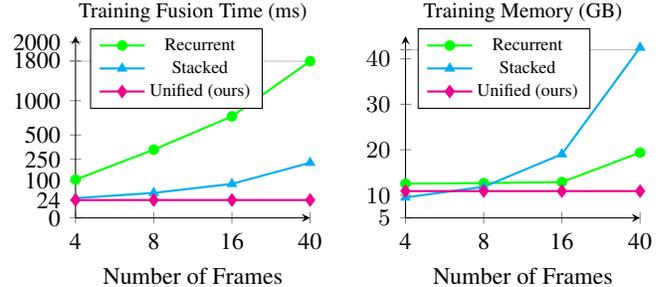

%% file: table/abl.tex
\begin{table*}[t]
\small
\setlength{\tabcolsep}{3.5pt}
\centering
\resizebox{\textwidth}{!}{
\begin{tabular}{@{}l|>{\colortable}c|ccccccccccccccccc@{}}
\toprule
Settings                    & mIoU  &\rotatebox{90}{others} &\rotatebox{90}{barrier} & \rotatebox{90}{bicycle} &  \rotatebox{90}{bus}   & \rotatebox{90}{car}   & \rotatebox{90}{const. veh.} & \rotatebox{90}{motorcycle} & \rotatebox{90}{pedestrian} & \rotatebox{90}{traffic cone} & \rotatebox{90}{trailer} & \rotatebox{90}{truck} & \rotatebox{90}{driv. surf.} & \rotatebox{90}{other flat} & \rotatebox{90}{sidewalk} & \rotatebox{90}{terrain} & \rotatebox{90}{manmade} & \rotatebox{90}{vegetation} \\
\midrule
No Temporal  & 37.39 & 12.17 & 44.83 & 25.73 & 42.61 & 47.97 & 23.16 & 25.17 & 25.77 & 26.72 & 31.31 & 34.89 & 78.83 & 41.42 & 49.06 & 52.22 & 39.07 & 34.61 \\
Mem. Attn. & 41.17 & 13.96 & 48.62 & 27.81 & 44.83 & 51.01 & 27.07 & 28.86 & 29.22 & 29.28 & 33.96 & 38.41 & 83.62 & 46.36 & 54.94 & 58.76 & 43.93 & 39.15 \\
Mem. Attn. + Dynamics & 41.73 & 14.03 & 48.46 & \textbf{28.02} & \textbf{47.23} & 52.10 & \textbf{27.72} & 28.81 & 29.43 & 29.97 & \textbf{36.28} & 38.81 & 84.00 & 45.66 & 55.45 & 59.24 & 44.43 & 39.75 \\
Mem. Attn. + Uncertainty & 41.85 & 14.35 & \textbf{50.31} & 27.48 & 46.26 & 51.93 & 27.62 & 28.68 & 29.28 & 29.82 & 34.60 & 38.92 & 84.09 & 47.32 & \textbf{56.23} & 59.54 & 45.00 & 39.98 \\
\rowcolor[RGB]{230,230,230} \method & \textbf{42.13} & \textbf{14.36} & 49.62 & 27.77 & 46.28 & \textbf{52.55} & 26.87 & \textbf{29.79} & \textbf{29.83} & \textbf{31.39} & 35.40 & \textbf{39.03} & \textbf{84.26} & \textbf{47.72} & 56.09 & \textbf{59.85} & \textbf{45.27} & \textbf{40.11} \\
\bottomrule
\end{tabular}
}
\caption{3D occupancy prediction results of \method with different design settings on Occ3D benchmark.}
\label{tab:abl}
\end{table*}

%% file: table/full_abl.tex
\begin{table}[ht]
\footnotesize
\begin{minipage}[t]{0.45\linewidth}\centering
  \begin{tabular}{cccc|c}
\toprule
 $\mathbf{c}$ & $\varepsilon$ & $\delta$ & $\mathbf{f}$ & mIoU \\
\midrule
    &       &            &      & 41.17     \\
 \cmark &       &            &      & 41.45     \\
 \cmark & \cmark &            &      &  41.73    \\
 \cmark & \cmark & \cmark &      &  41.85    \\
     &       &            & \cmark &  41.73    \\
 \cmark & \cmark & \cmark & \cmark &  42.13   \\
\bottomrule
\end{tabular}
\end{minipage} \hfill
\begin{minipage}[t]{0.5\linewidth}
\vspace{-40pt}
\caption{Ablation study on different sub-components of \method on Occ3D benchmark. $\mathbf{c}$, $\varepsilon$, $\delta$, and $\mathbf{f}$ correspond to the historical class activation, feature similarity, averaged log variance, and occupancy flow.}
\vspace{-40pt}
\label{tab:full_abl}
\end{minipage}
\end{table}

%% file: table/frame_curve_sep.tex
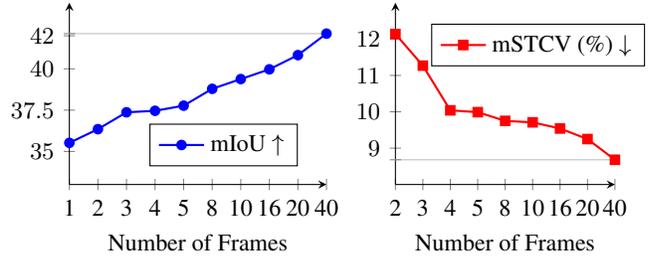
\begin{figure}[t!]
    \centering
    % \vspace{-1em}
    % First subplot for mIoU
    \begin{tikzpicture}
    \hspace{-2.0em}
        \begin{axis}[
            width=5cm,
            height=4cm,
            axis lines=left,
            xlabel={Number of Frames},
            ylabel style={yshift=-15pt},
            symbolic x coords={1, 2, 3, 4, 5, 8, 10, 16, 20, 40},
            xtick=data,
            ymin=33, ymax=44,
            ytick={30, 35, 37.5, 40, 42, 45},
            legend style={at={(0.6,0.10)}, anchor=south, font=\small},
            label style={font=\small},
            tick label style={font=\small},
            extra y ticks={42.13},
            extra y tick labels={},
            extra y tick style={grid=major},
        ]
            % Plot mIoU
            \addplot[
                color=blue,
                mark=*,
                mark options={scale=0.8},
                thick
            ] coordinates {
                (1, 35.52)
                (2, 36.35)
                (3, 37.37)
                (4, 37.46)
                (5, 37.77)
                (8, 38.79)
                (10, 39.38)
                (16, 39.97)
                (20, 40.83)
                (40, 42.13)
            };
            \addlegendentry{mIoU $\uparrow$}
        \end{axis}
    \end{tikzpicture}
    % Space between subplots
    % \hspace{0.05\columnwidth}
    % Second subplot for m\metric
    \hspace{-2.5em}
    \begin{tikzpicture}
        \begin{axis}[
            width=4.5cm,
            height=4cm,
            axis lines=left,
            xlabel={Number of Frames},
            ylabel style={yshift=-15pt},
            symbolic x coords={2, 3, 4, 5, 8, 10, 16, 20, 40},
            xtick=data,
            ymin=8, ymax=13,
            ytick={5, 9, 10, 11, 12, 15, 20, 25},
            legend style={at={(0.65,0.65)}, anchor=south, font=\small},
            label style={font=\small},
            tick label style={font=\small},
            extra y ticks={8.68},
            extra y tick labels={},
            extra y tick style={grid=major},
        ]
            % Plot m\metric
            \addplot[
                color=red,
                mark=square*,
                mark options={scale=0.8},
                thick
            ] coordinates {
                (2, 12.14)
                (3, 11.27)
                (4, 10.04)
                (5, 9.99)
                (8, 9.75)
                (10, 9.71)
                (16, 9.54)
                (20, 9.25)
                (40, 8.68)
            };
            \addlegendentry{m\metric (\%) $\downarrow$}
        \end{axis}
        % \hspace{6em}
    \end{tikzpicture}
    % \hspace{-2em}
    \vspace{-1em}
    \caption{Effect of number of frames incorporated during inference on the precision and robustness of occupancy prediction. We observe a strong correlation between the number of incorporated frames and the studied metrics.
}
    \vspace{-1em}
    \label{tab:frames}
\end{figure}

%% file: suppl.tex
\clearpage
% \setcounter{page}{1}
% \maketitlesupplementary
\section*{Appendix}
\appendix

\def\thesection{\Alph{section}}
\setcounter{section}{0} % Reset section counter to 0
\setcounter{table}{0}
\setcounter{figure}{0}

\section{Implementation Details}

In this section, we provide implementation details of the proposed \method and experimental setup.

\subsection{Spatiotemporal Memory}

Our spatiotemporal memory $\mathbf{M}$ has a channel size of $C_G = 101$, which includes historical representation with 80 channels, historical class activations $\mathbf{c}$ with 18 channels, occupancy flow with 2 channels, and averaged log variance with 1 channel. 

The spatial dimension $H_G, W_G$ is determined at the beginning of each scene according to the ego-motion, and $Z_G = Z = 8$. 
From an analytical perspective, recurrent-based or stack-based approaches require $O(kHWZ)$ memory to store the historical frame representations, where $k$ is the number of timesteps used (e.g., $k=40$ for 20s at FPS 2, or $k=200$ at FPS 10). In contrast, our temporal modeling needs $O\left((1+\Delta)HWZ\right)$ memory, where $\Delta$ represents the relative volume change and it is much smaller than $k$. For example, in all nuScenes samples, 
$\Delta_{\text{mean}}=3.25, \Delta_{\text{max}}=16.75$, and $\Delta_{\text{min}}=0.56$
over a 20-second duration and it is FPS-\textit{agnostic}. In other words, our method requires, on average, only $\frac{1}{10}$ of the memory compared to queue-based approaches and at most $\frac{1}{2}$ of the memory in the worst case.

\subsection{Feature Sampling}

The feature sampling operation $\chi[\cdot]$ is used to extract the ego vehicle-centered representation at timestamp $t$ from our spatiotemporal memory $\mathbf{M}_t$ given the ego vehicle pose $T_t$. For the voxel grid $\mathcal{G} = \left\{ \mathbf{p} \in \mathbb{R}^3 \mid \mathbf{p} = (x, y, z), x \in [0, W), y \in [0, H), z \in [0, D) \right\}$ with the ego vehicle in the center, we use the ego pose matrix $T_t \in \mathbb{R}^{4 \times 4}$ provided in the dataset for transformation. We first express $\mathbf{p}$ in homogeneous coordinates as
% \vspace{-0.8em}
% {\footnotesize
\begin{equation}
    \tilde{\mathbf{p}} = \begin{bmatrix} x & y & z & 1 \end{bmatrix}^T.
\end{equation} % }
We then transform the ego vehicle-centered grid into our spatiotemporal memory coordinate system according to
% \vspace{-0.7em}
% {\footnotesize
\begin{equation}
    \mathcal{G}_\mathcal{M} = \left\{ \left[ \mathbf{T} \cdot \begin{bmatrix} x & y & z & 1 \end{bmatrix}^T \right]_{1:3} \mid (x, y, z) \in \mathcal{G} \right\}.
\end{equation}% }
We use this transformed grid for sampling via trilinear interpolation.
The sampled representation is then used for downstream occupancy prediction.

When updating the spatiotemporal memory with ego vehicle-centered representation or temporal attributes, the process follows
\begin{equation}
    \chi[\mathbf{M}_{t+1}\langle \text{value} \rangle, T_t] = \text{value}_t.
\end{equation}
We first determine the corresponding region of interest (RoI) in scene-centered coordinates. For each voxel in the RoI, its location is transformed into ego vehicle-centered coordinates using the inverse ego pose $T_t^{\text{inv}}$. Then we perform grid sampling with bilinear interpolation to ensure accurate feature retrieval. This process ensures that each historically traversed location in the spatiotemporal memory is updated, thereby mitigating misalignments caused by coordinate transformations.

\subsection{Memory Attention}

Our memory attention consists of 3 temporal self-attention (TSA) layers \cite{li2022bevformer}, each performing operations in the sequence of \textit{self-attention}, \textit{normalization}, \textit{feedforward}, and \textit{normalization}. A 3D learnable position embedding is added to the query. For the deformable attention in the \textit{self-attention} operation, we use four sampling points for each reference point corresponding to the query. 

The network used to encode temporal attributes is a 4-layer Multi-Layer Perceptron (MLP) with hidden sizes 64, 32, 16, and 1, respectively. The encoded temporal attributes $u$ and occupancy flow $\mathbf{f}$ are shared across all three TSA layers.

Our method applied a three-layer convolutional network on ego vehicle-centered representation for occupancy flow prediction. It achieves an mAVE of 0.618, which is on par with existing methods \cite{OpenDriveLab2024} and ensures its reliability and efficiency.

\begin{figure}[t]
\renewcommand{\thefigure}{A}
    \centering
    \includegraphics[width=0.48\textwidth]{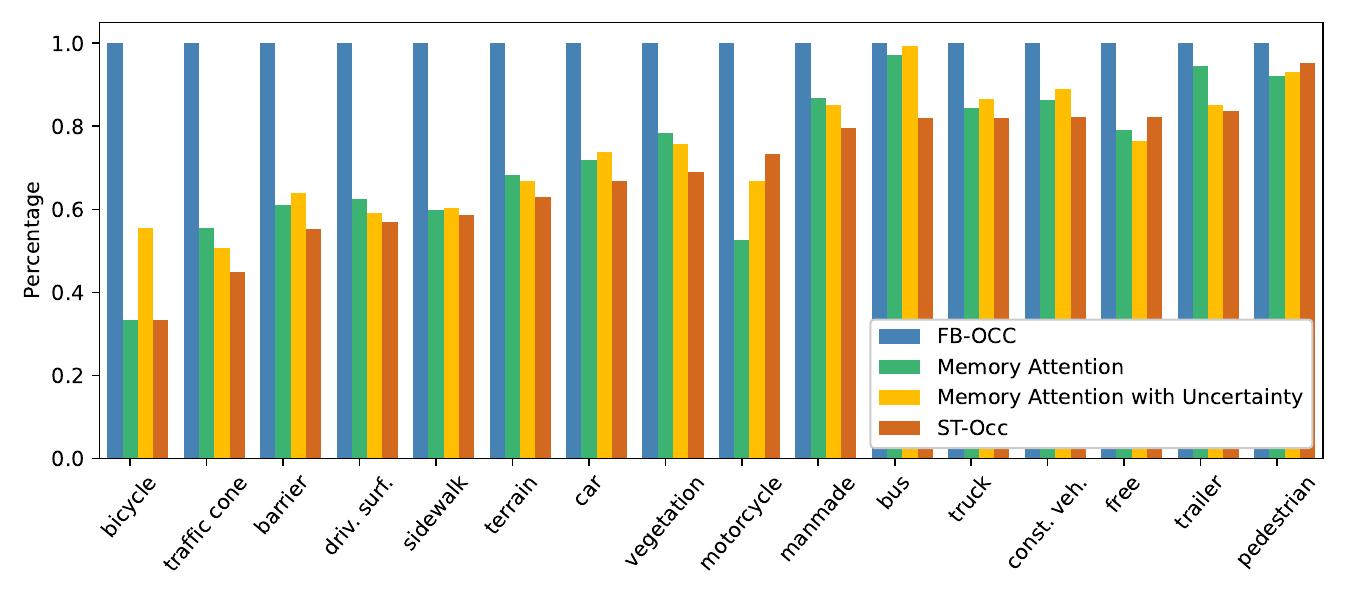}
   \caption{The temporal consistency evaluation results of each class.  \textcolor{SteelBlue}{FB-OCC} is set as the baseline, followed by different settings of our method. We compute the relative m\metric reduction with respect to baseline (the lower the better). Thus the results  for \textcolor{SteelBlue}{FB-OCC} are all 1s.}
   \label{fig:temp_cons_dist}
   \vspace{-1em}
\end{figure}

\section{Temporal Consistency}

Our memory attention, equipped with uncertainty awareness, also results in better temporal consistency of the occupancy prediction. Results in \cref{tab:temporal_cons} show that uncertainty awareness contributes an additional 2\% improvement in temporal consistency by reducing m\metric. This highlights the effectiveness of uncertainty modeling in mitigating noise accumulation.

To further demonstrate the effectiveness of our design in reducing temporal inconsistencies in occupancy prediction, we evaluate temporal consistency across individual classes in \cref{fig:temp_cons_dist}. Results reveal a 40\% reduction in temporal inconsistency for static classes with our memory attention. With the uncertainty and dynamic awareness incorporated, our \method can further reduce inconsistency for certain classes. The decrease in temporal inconsistency is consistent with the increase in occupancy prediction regarding various object classes.  Notably, classes such as \textit{barrier}, \textit{traffic cone}, and \textit{drivable surface}, which exhibit lower temporal inconsistency, also achieve higher occupancy prediction accuracy than the baseline. These findings not only verify the effectiveness of our method but also highlight the importance of reducing temporal inconsistency in occupancy prediction, thereby providing more reliable and robust predictions for downstream tasks.

\section{Historical Occupancy Prediction}

To evaluate the models' performance in preserving and utilizing historical information, we extend the original occupancy prediction evaluation scope while maintaining the mIoU metric unchanged. During the evaluation, we included not only the visible voxels in the current timestamp but also any invariant voxels visible in the previous timestamp. Voxels corresponding to dynamic objects in historical frames are excluded to ensure evaluation consistency. Furthermore, these historically visible invariant voxels can be incorporated during training to enhance occupancy learning.

The results in \cref{tab:supp_geval} demonstrate that including historically visible voxels in the evaluation leads to a lower mIoU score than the original setting, as accurately predicting these voxels is inherently more challenging. Despite this harder evaluation, our proposed \method outperforms the FB-OCC by a margin of 5\%. Additionally, when historically visible voxels are incorporated during training, our method achieves the highest performance on the extended evaluation scope. The observed performance improvement in FB-OCC indicates that training with historically visible voxels benefits the occupancy prediction.

\input{table/supp_geval}

\section{Training Analysis}

\cref{fig:training} illustrates the training curve of our method compared with the baseline. While our method slightly underperforms FB-OCC \cite{li2023fb} during the initial epochs, this is attributed to the more complex network architecture deployed for temporal fusion. However, our method demonstrates faster convergence and performs comparable to FB-OCC in nearly half the training epochs. In the end, the \method delivers impressive performance with significant performance improvements.

\input{table/supp_training_curve}

\section{Visualization}

\subsection{Uncertainty}

\begin{figure}[t]
\renewcommand{\thefigure}{C}
    \centering
    \resizebox{0.9\linewidth}{!}{ % Adjust 0.9 to desired width
\begin{minipage}{\linewidth}
    \begin{subfigure}{0.48\linewidth}
        \centering
        \includegraphics[width=\linewidth, trim=200 0 0 280, clip]{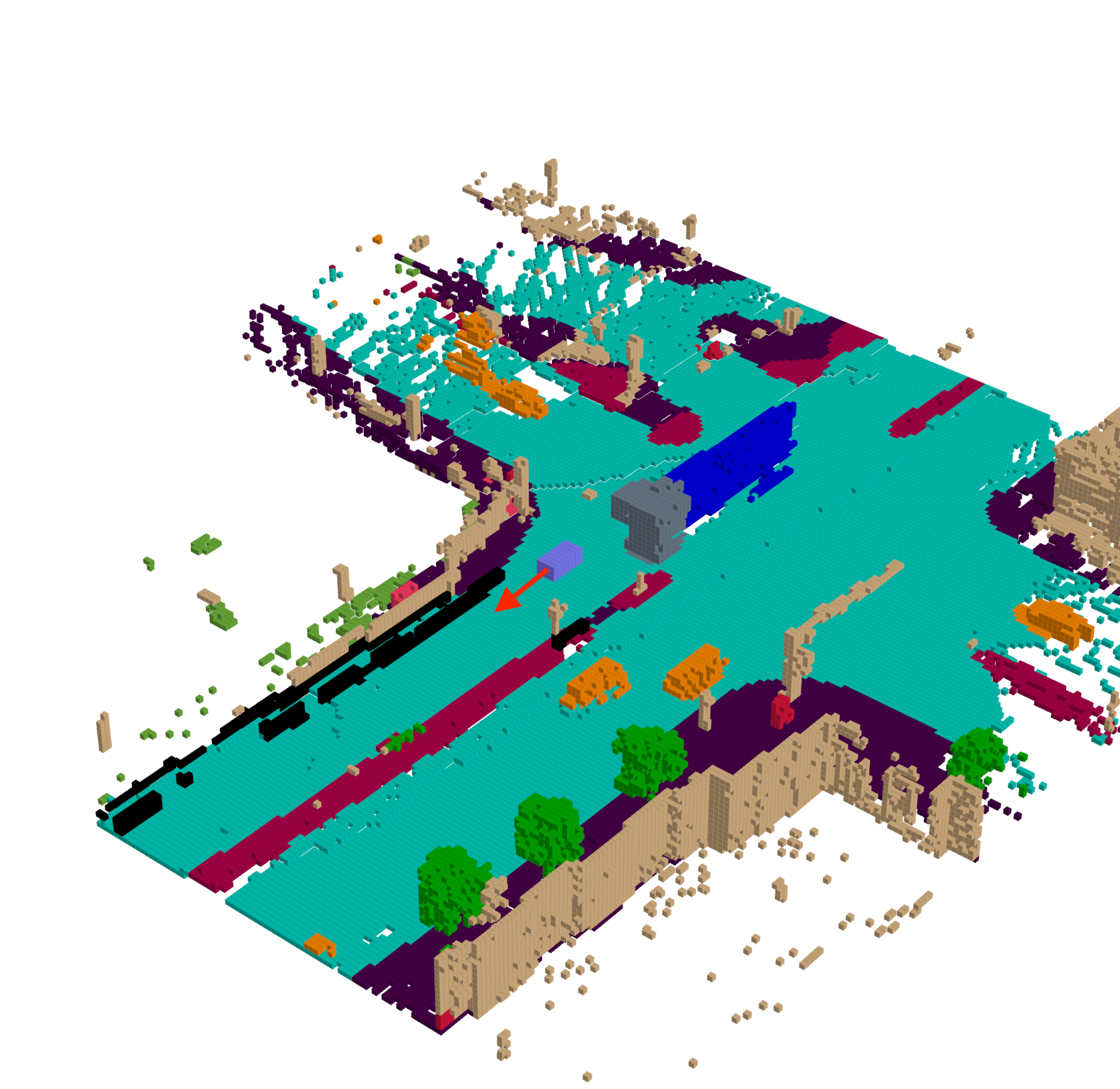}
        \caption{\method Prediction (\textcolor{egopurple}{Ego Veh.})}
        \label{fig:uc_pred}
    \end{subfigure}
    \hfill
    \begin{subfigure}{0.48\linewidth}
        \centering
        \includegraphics[width=\linewidth, trim=200 0 0 280, clip]{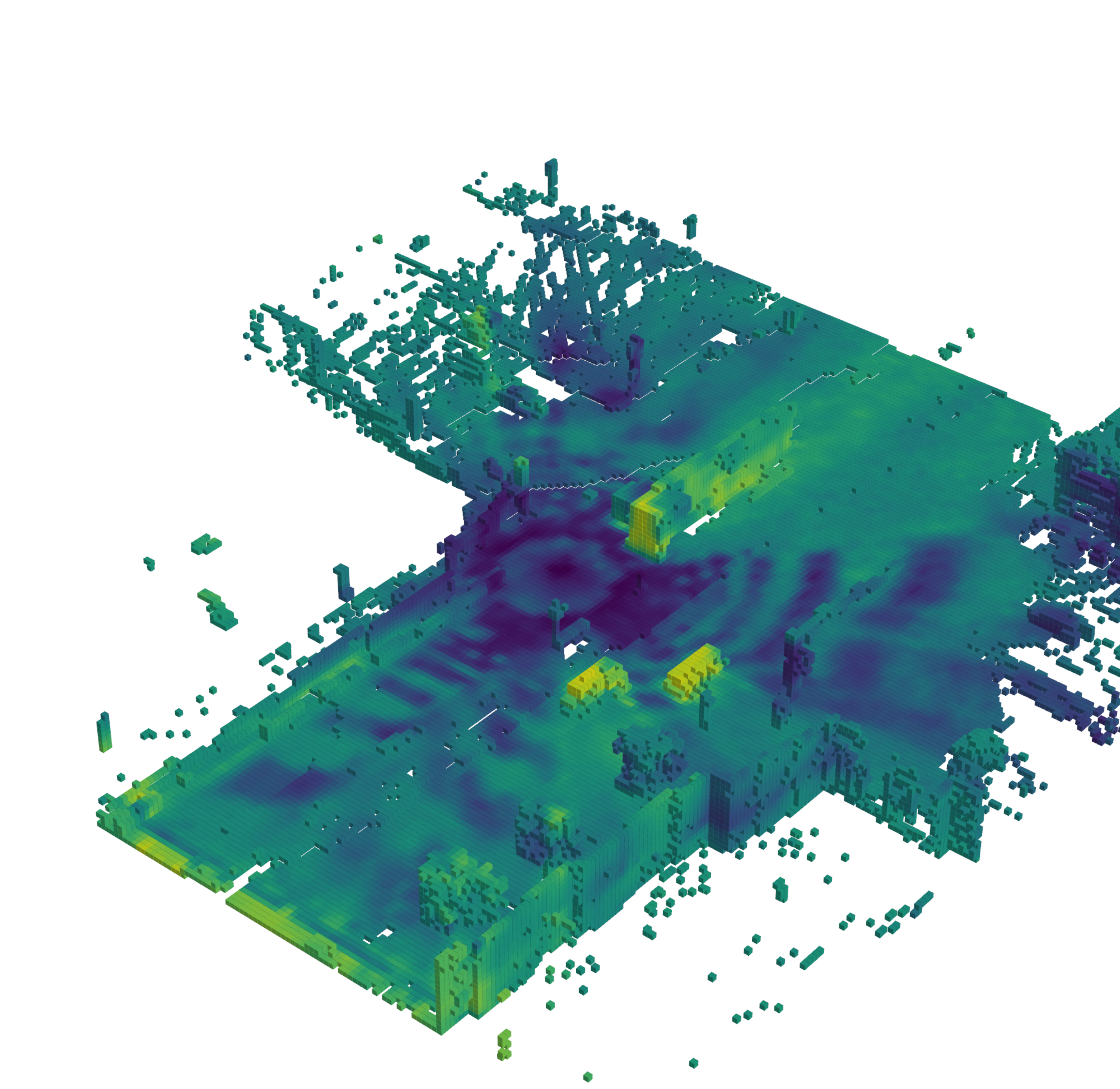}
        \caption{Uncertainty from \method}
        \label{fig:uc_uc}
    \end{subfigure}
\end{minipage}
}
    % \vspace{-1.em}
    \caption{Visualization of uncertainty in \method.}
    \label{fig:uncertainty_vis}
    % \vspace{-0.8em}
\end{figure}

We visualize uncertainty estimated by our method in \cref{fig:uncertainty_vis}, where dynamic, occluded, and unobserved voxels have higher uncertainty while observed regions show lower uncertainty.

\subsection{Occupancy Prediction}
We present visualizations of ego vehicle-centered and scene-level occupancy prediction done by the proposed method on additional large-scale scenes in \cref{fig:supp_vis2}. Our method can produce precise occupancy predictions and construct a comprehensive scene representation.

The minor inconsistencies observed between ego vehicle-centered and scene-level prediction (particularly evident in frame 20 of \cref{fig:scene103} and \cref{fig:scene523}) can be attributed to two factors: 1) Continuous Updates. The RoI regarding each frame in the spatiotemporal memory is updated incrementally by subsequent frames with additional observations. 2) Dynamic Instances. Our pipeline does not incorporate explicit dynamic object masking in the spatiotemporal memory. Instead, we rely on memory attention with uncertainty and dynamic awareness to handle dynamic voxels implicitly.

% \vspace{2em}

% \FloatBarrier
\renewcommand{\thefigure}{D}
\renewcommand{\thesubfigure}{\roman{subfigure}}
\begin{figure*}[!h]
    \centering
    % First subfigure
    \begin{subfigure}[b]{\textwidth}
        \centering
        \includegraphics[width=1.0\textwidth, trim=0 60 0 40, clip]{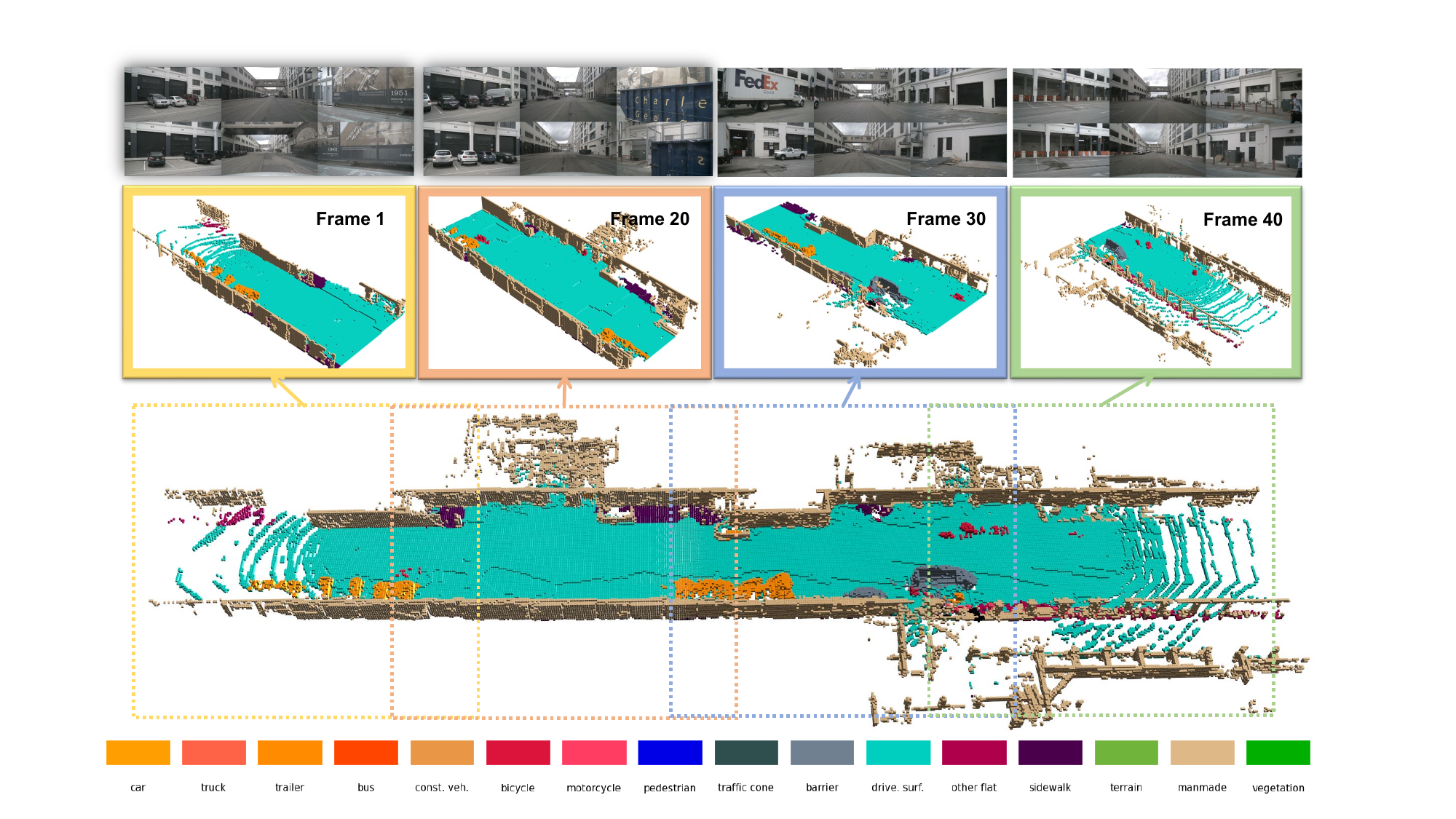} % Replace with your image
        \caption{\textit{Scene-0092}}
        \label{fig:scene92}
    \end{subfigure}
    
    \vspace{1cm} % Adjust spacing as needed
    
    % Second subfigure
    \begin{subfigure}[b]{\textwidth}
        \centering
        \includegraphics[width=1.0\textwidth, trim=0 0 0 0, clip]{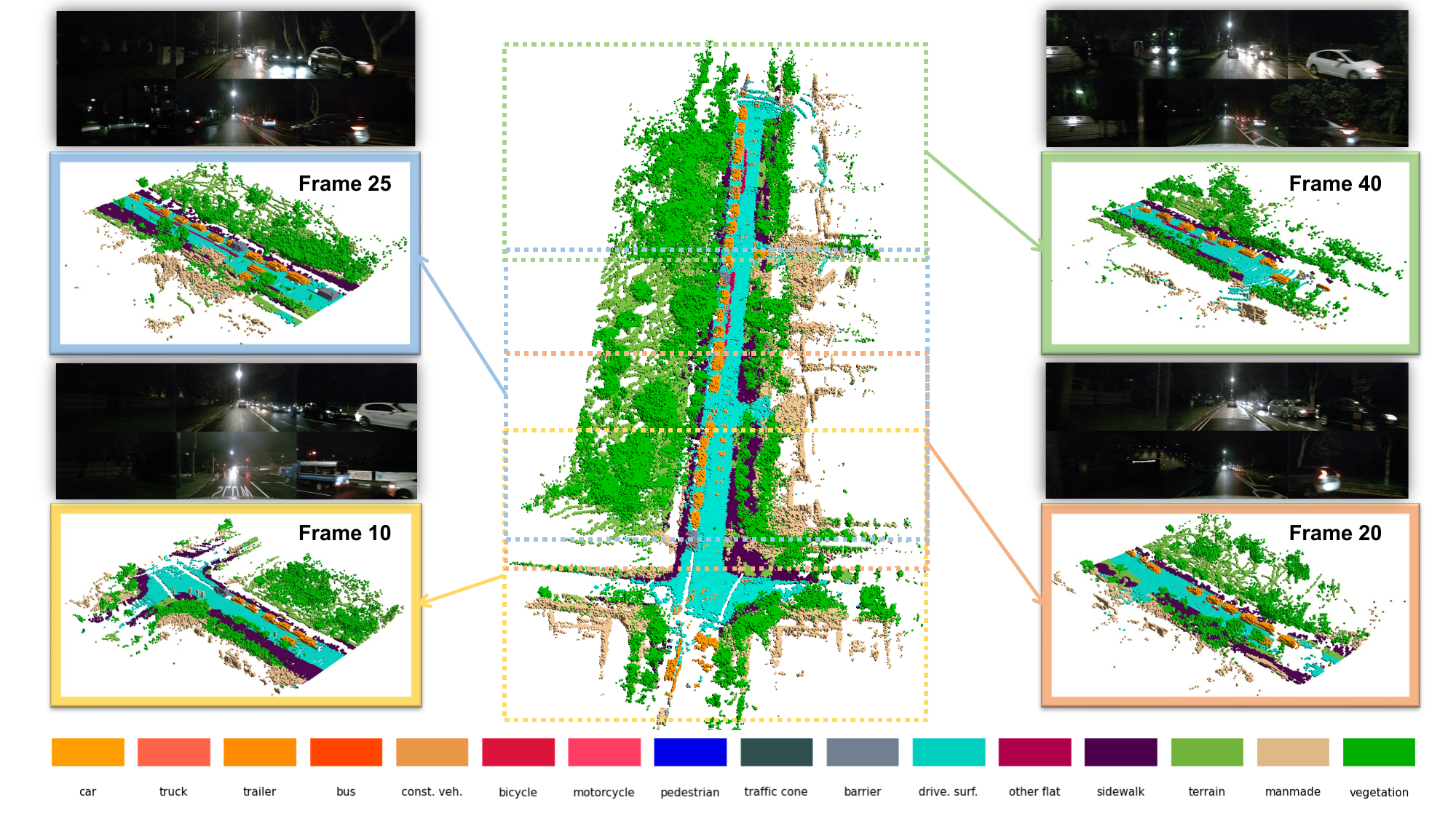} % Replace with your image
        \caption{\textit{Scene-1069}}
        \label{fig:scene1069}
    \end{subfigure}
    \label{fig:supp_vis}
\end{figure*}

\begin{figure*}[h]
    \ContinuedFloat
    \centering
    % First subfigure
    \begin{subfigure}[b]{\textwidth}
        \centering
        \includegraphics[width=1.0\textwidth, trim=0 55 0 30, clip]{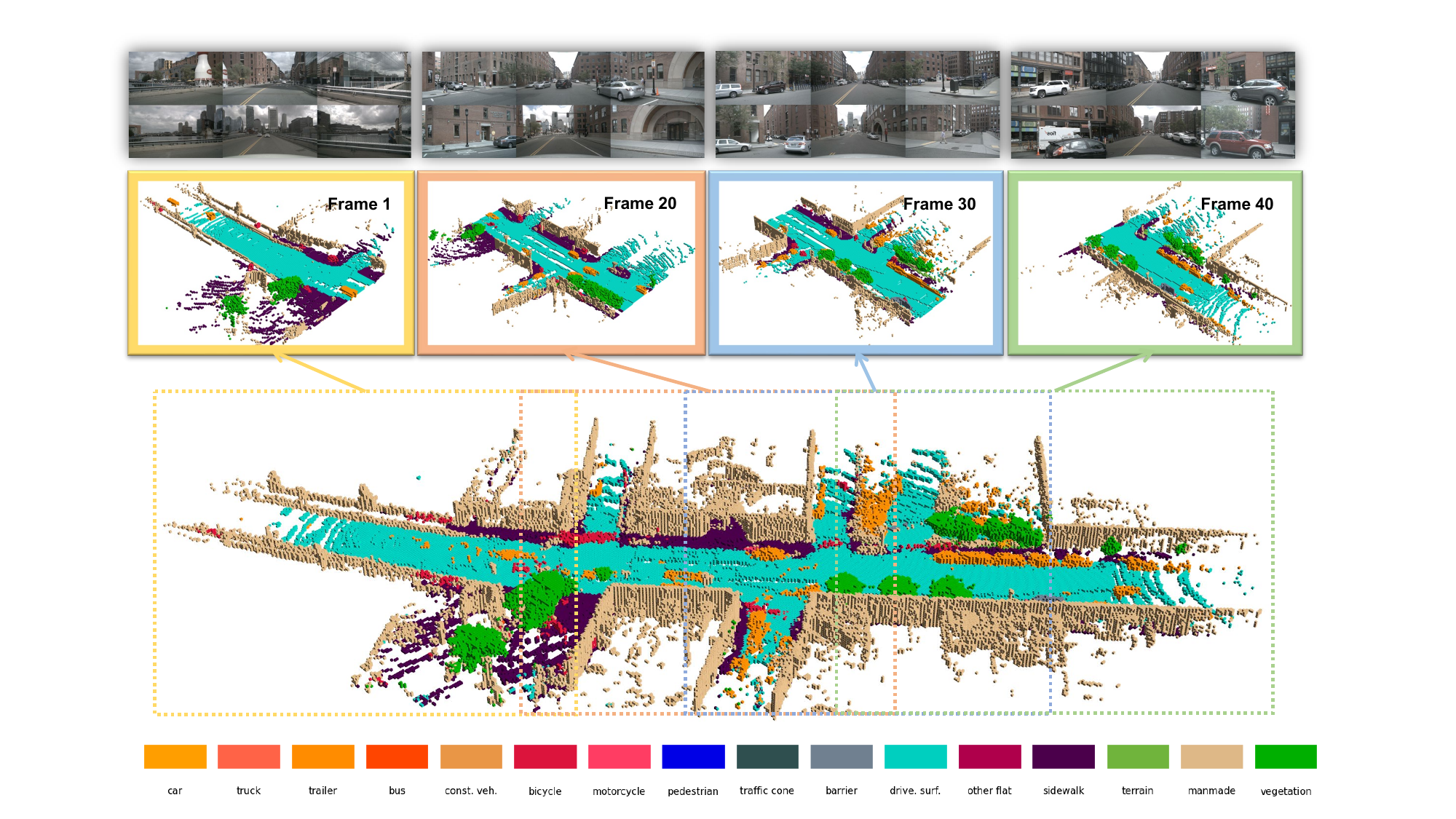} % Replace with your image
        \caption{\textit{Scene-0103}}
        \label{fig:scene103}
    \end{subfigure}
    
    \vspace{1cm} % Adjust spacing as needed
    
    % Second subfigure
    \begin{subfigure}[b]{\textwidth}
        \centering
        \includegraphics[width=1.0\textwidth, trim=0 0 0 30, clip]{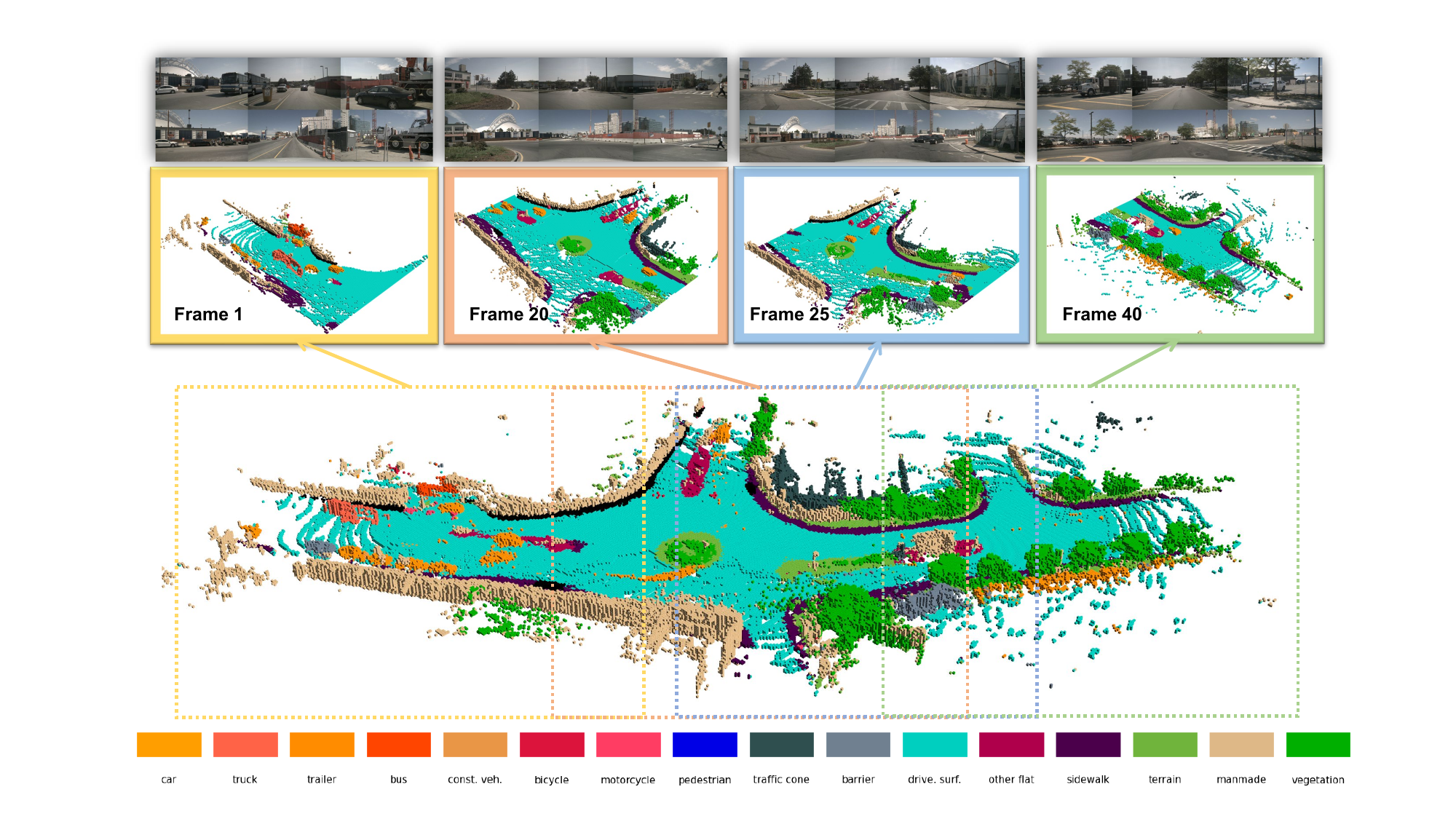} % Replace with your image
        \caption{\textit{Scene-0523}}
        \label{fig:scene523}
    \end{subfigure}
    \caption{Visualization of occupancy prediction results from \method across four representative scenes. The ego vehicle-centered predictions at different timestamps are displayed within solid-line boxes, with the corresponding input RGB images provided above. Each scene also includes the scene-level occupancy prediction derived from our spatiotemporal memory aggregated over all frames.}
    \label{fig:supp_vis2}
\end{figure*}

%% file: table/supp_geval.tex
\begin{table}[t]
\small
\renewcommand{\thetable}{A}
\centering
\begin{tabular}{@{}l|cc}
\toprule
Method                            & mIoU & mIoU$^{\dagger}$  \\ 
\midrule
FB-OCC                            & 39.11                   & 33.71                    \\
\method                  & \textbf{42.13}                    & 35.34                    \\
\midrule
FB-OCC$^{\ddagger}$ & 40.06                    & 35.78                  \\
\method$^{\ddagger}$           & 41.62                    & \textbf{36.96}                   \\ 
\bottomrule
\end{tabular}
\caption{Historical occupancy prediction results on the extended Occ3D benchmark. $^{\dagger}$ denotes evaluation with historically visible voxels included. $^{\ddagger}$ incorporates historically visible voxels during training.}
\label{tab:supp_geval}
\end{table}

%% file: table/supp_training_curve.tex
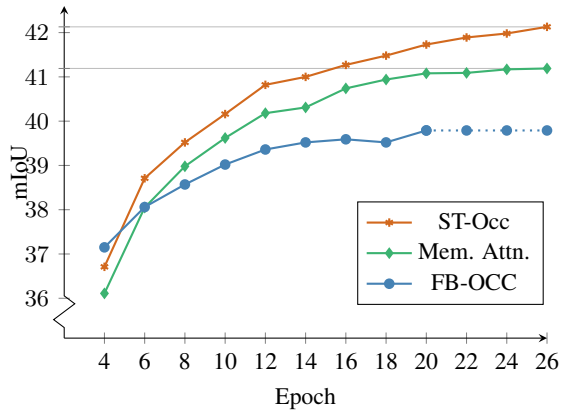
\begin{figure}[t!]
\renewcommand{\thefigure}{B}
    \centering
\begin{tikzpicture}
\hspace{-1.0em}
    \begin{axis}[
        axis lines=left,
        width=8cm, % Adjust the width of the plot
        height=6cm, % Adjust the height of the plot
        xlabel={Epoch}, % Label for the x-axis
        ylabel={mIoU}, % Label for the y-axis
        ylabel style={yshift=-12pt},
        ymin=35.1, ymax=42.6,
        xmin=2, xmax=26,
        axis y discontinuity=crunch,
        ytick distance = 1,
        legend style={at={(0.8,0.10)}, anchor=south, font=\small},
            label style={font=\small},
            tick label style={font=\small},
        % grid=both, % Add grid
        % minor tick num=1, % Add minor ticks
        % major grid style={line width=0.2pt,draw=gray!50},
        % minor grid style={line width=0.1pt,draw=gray!20},
        xtick={4,6,8,10,12,14,16,18,20,22,24,26}, % Custom x-axis ticks
        xticklabels={4,6,8,10,12,14,16,18,20,22,24,26}, % Labels for the x-axis ticks
        extra y ticks={42.13, 41.19},
            extra y tick labels={},
            extra y tick style={grid=major},
    ]

        % Series 3
        \addplot[
            % smooth,
            thick,
            Terracotta,
            mark=asterisk,
            mark options={scale=0.8},
        ] table {
            4 36.71 
            6 38.71 
            8 39.52 
            10 40.16 
            12 40.82 
            14 41.0 
            16 41.27 
            18 41.48 
            20 41.73 
            22 41.89
            24 41.98
            26 42.13
        };
        \addlegendentry{\method}
        
        % Series 2
        \addplot[
            % smooth,
            thick,
            TealGreen,
            mark=diamond*,
            mark options={scale=0.8},
        ] table {
            4 36.11 
            6 38.05 
            8 38.98 
            10 39.62
            12 40.18
            14 40.31
            16 40.74
            18 40.94
            20 41.08
            22 41.09
            24 41.17
            26 41.19
            
        };
        \addlegendentry{Mem. Attn.}
        
        % Series 1
        \addplot[
            % smooth, % Creates a curve
            thick, % Line thickness
            SteelBlue, % Line color
            mark=*,
            mark options={scale=0.8},
        ] table {
            % 2 27.07
            4 37.15 
            6 38.06 
            8 38.57 
            10 39.02
            12 39.36
            14 39.52
            16 39.59
            18 39.52
            20 39.79
        };
        \addlegendentry{FB-OCC}

        \addplot[
            % smooth, % Creates a curve
            thick, % Line thickness
            dotted,
            SteelBlue, % Line color
            mark=*,
            mark options={solid, scale=0.8},
        ] table {
            20 39.79
            22 39.79
            24 39.79
            26 39.79
        };
    \end{axis}

\end{tikzpicture}
\caption{Training curve of our \method and Memory Attention compared to FB-OCC. Our approach demonstrates faster convergence while achieving impressive performance.}
\label{fig:training}
\end{figure}